%% file: main.tex
\documentclass[10pt,twocolumn,letterpaper]{article}

\usepackage[pagenumbers]{cvpr} 

\input{preamble}
\definecolor{cvprblue}{rgb}{0.21,0.49,0.74}
\usepackage[pagebackref,breaklinks,colorlinks,allcolors=cvprblue]{hyperref}
\usepackage{siunitx}
\usepackage{xspace}
\usepackage{xcolor}
\usepackage[T1]{fontenc}
\usepackage{newtxtext,newtxmath}
\usepackage[accsupp]{axessibility}  
\usepackage{color,soul}

\definecolor{BetterGreen}{RGB}{34,139,34}
\definecolor{BetterRed}{RGB}{178,34,34}
\definecolor{bestyellow}{RGB}{255, 245, 200}
\definecolor{BetterOrange}{RGB}{255,165,0}

\newcommand{\movierecapsqa}{\textsc{MovieRecapsQA}\xspace}

\title{\movierecapsqa: A Multimodal Open-Ended Video Question-Answering Benchmark}

\author{Shaden Shaar$^*$ \and Bradon Thymes$^*$ \and Sirawut Chaixanien \and Claire Cardie \and Bharath Hariharan\\
Cornell University\\
{\tt\small \{ss2753, bmt63\}@cornell.edu}
}

\begin{document}
\twocolumn[{
\renewcommand\twocolumn[1][]{#1}
\maketitle
\small
\begin{center}
    \vspace{-3mm}
    \includegraphics[width=.9\textwidth]{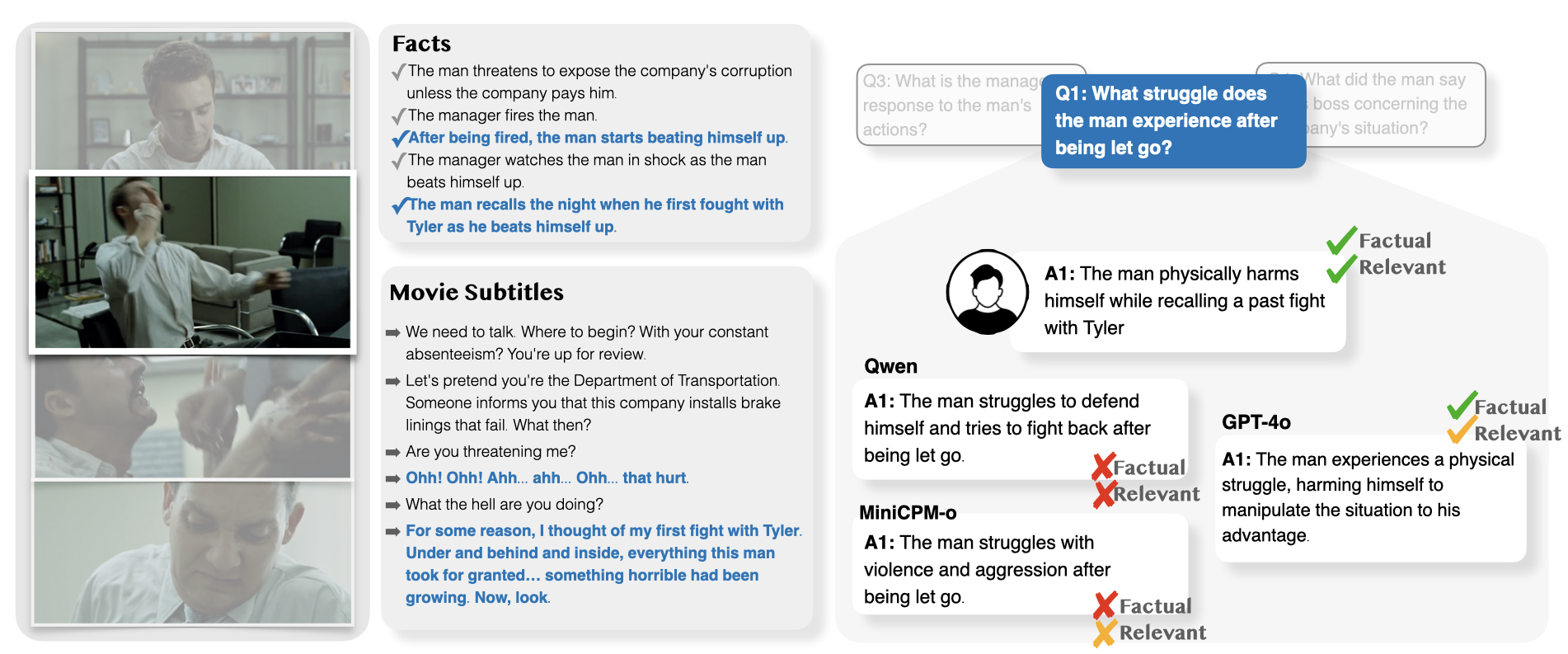}
    \vspace{-3mm}
    \captionsetup{hypcap=false}
    \captionof{figure}{%
        \textbf{\movierecapsqa Benchmark.}
        An example (movie \textit{Fight Club}) from our benchmark illustrating how MLLMs answer questions using recap-video frames and aligned movie subtitles. Q1 was constructed from Facts (3) and (5), and answering it requires integrating visual cues (Frame 2) with supporting subtitle evidence (Lines 4 and 6). We show the human answer alongside model outputs, evaluated for relevance and factuality on a 0–5 fact-grounded scale, with colors indicating quality from lowest to highest: {\color{BetterRed}{\sffamily X}}, {\color{BetterOrange}{\sffamily X}}, {\color{BetterOrange}\checkmark}, {\color{BetterGreen}\checkmark}.%
        \label{fig:teaser}
    }
    \vspace{3mm}
\end{center}
}]
\begingroup
\hypersetup{hypertexnames=false}
\def\thefootnote{*}
\footnotetext{These authors contributed equally to this work}
\endgroup
\renewcommand{\thefootnote}{\arabic{footnote}}

\input{sections/0.abstract}
\input{sections/1.introduction}
\input{tables/table-compare}
\input{sections/2.related-work}

\input{sections/3.dataset}
\input{sections/4.evaluation-metrics}

\input{sections/6.experiment}
\input{sections/7.analysis}
\input{sections/8.conclusion}

{
    \small
    \bibliographystyle{ieeenat_fullname}
    \bibliography{main}
}

\appendix
\clearpage
\maketitlesupplementary
\setcounter{page}{1}
\renewcommand{\thefigure}{S\arabic{figure}}
\setcounter{table}{0}
\renewcommand{\thetable}{S\arabic{table}}

\input{supplementary/A.copyright}

\input{supplementary/B-C.dataset-details}

\input{supplementary/D.problem-setting}

\input{supplementary/E.question-categorization-and-modality}
\input{supplementary/F.evaluation-prompts}

\input{supplementary/G.coherence-study}

\input{supplementary/H.detailed-ablation-experiments}

\input{supplementary/Z.example-questions}
\input{supplementary/Z.examples-claims-facts}

\input{supplementary/Z.all_prompts}

\end{document}

%% file: preamble.tex
\usepackage{xcolor}
\usepackage[table]{xcolor}

\usepackage{listings}
\usepackage{array}

\usepackage{colortbl}
\usepackage{xspace}

\usepackage{tabularx}
\usepackage{siunitx} 
\usepackage{makecell}

%% file: sections/0.abstract.tex
\begin{abstract}
Understanding real-world videos such as movies requires integrating visual and dialogue cues. Yet existing VideoQA benchmarks struggle to capture this multimodal reasoning and, given the difficulty of evaluating free-form answers,  largely resort to simple multiple choice questions.
We introduce a novel open-ended multimodal VideoQA benchmark, \movierecapsqa, created using movie recap videos -- a distinctive type of YouTube content that summarizes a film via a voiceover description 
of key clips from the movie (recap video).
From the transcribed voiceover (recap summary) of 60 recap videos, 
we generate $\approx$8.2K questions along with 
the necessary ``facts'' expected in each answer; the former facilitates the creation of questions that require mutimodal reasoning and the latter allow the construction of a reference-free evaluation metric that can be applied to open-ended responses. 
To our knowledge, this is the first reference-free open-ended VideoQA benchmark.
The benchmark allows each question to be evaluated in different input video settings: given 
(a) the full-length 
movie, 
(b) the full ($\approx$11 min) recap video (visual only), (c) $\approx$14 min of aligned movie scenes, i.e, movie scenes relevant to the question, and (d) $\approx$1.2 min of aligned recap video scenes. 
In all cases, the text of any associated movie dialogue is provided.
Each question is categorized by the modality required to answer it---visual, dialogue, or both---enabling fine-grained evaluation of multimodal capabilities.
We benchmark (setting (d)) seven state-of-the-art MLLMs  
and find that 
(i) only our reference-free metric produces meaningful human-aligned model separation; 
(ii) vision-centric questions yield the lowest scores across all models;  
(iii) removing visual input often \textit{improves} 
model factuality;
and (iv) the primary bottleneck is visual perception, not visual reasoning.
\end{abstract}

%% file: sections/1.introduction.tex
\section{Introduction}
\label{sec:intro}

Consider the prospect of a system that can watch a movie like \emph{Fight Club} (Figure~\ref{fig:teaser}) and then answer questions about the movie.
To answer a question like ``What struggle does the man experience after being let go?", the system would need to listen to the dialogue to determine when the person is being  fired (i.e., ``let go"), and then look at the corresponding video sequence to see how the person reacts.
The need for multimodal reasoning across video and dialogue is not limited to the context of movie understanding; it generalizes to real world tasks, such as robot reasoning over visual and linguistic modalities. 

Most existing Video Question Answering (VideoQA) benchmarks fall short of capturing the complexity of such multimodal understanding.
Many benchmarks focus on a single modality (e.g., solely visual or solely audio) \cite{mangalam2023egoschemadiagnosticbenchmarklongform} and often feature simple thematic questions and relatively short video clips.
Even recent work that explicitly focuses on long videos and multimodal reasoning uses simpler multiple-choice questions \cite{rawal_cinepile_2024, wu2024longvideobenchbenchmarklongcontextinterleaved, hu2025videommmu}. 
In contrast to open-ended question answering, 
multiple choice questions offer shortcuts for answering the question without understanding the video.
Thus, current benchmarks provide only limited understanding of how well models generalize to real-world content, such as movies and real-life social contexts, where understanding the temporal dynamics and fine-grained character interactions require complex multimodal reasoning.

A key reason for the lack of better benchmarks is the sheer difficulty of acquiring complex questions and answers --- careful manual construction of high-quality QA pairs is time-consuming.  Evaluating model answers in this open-ended setting is also hard: gold standard answers are not typically
single words or simple phrases, and comparing them to model-generated answers is subjective.  As a result, evaluation strategies rely on
the use of word-based semantic similarity metrics (e.g., ROUGE) that
have been shown to have a low correlation with human judgments on factuality and quality \cite{kamalloo-etal-2023-evaluating, pagnoni-etal-2021-understanding, fabbri2021summevalreevaluatingsummarizationevaluation}.

The evaluation challenge is further magnified in the video domain. In text-based question answering, 
using LLMs as a judge has been proposed to verify an answer's factuality and relevance with respect to a source text. 
However, this approach is not directly transferable to VideoQA: using the full video as ``context'' for a  judge is computationally expensive and imprecise. The model would have to parse complex, multimodal signals just to find the relevant information for answer grounding. Consequently, existing attempts to use LLM judges for VideoQA often fall back on the reference-answer paradigm \cite{ma2025videoevalprorobustrealisticlong, cao2025videosimpleqafactualityevaluation}.

We address these challenges by tapping into a novel data 
source: \emph{movie recaps}. A movie recap is a distinctive type of YouTube content that provides a short ($\sim$10 min) summary of a full-length film via a voiceover description 
of key visual-only clips from the movie. We will refer to the clips as the \textbf{RecapVideo} and the transcribed voiceover as the \textbf{RecapSummary}.
Importantly, the RecapSummary can be automatically mined to extract facts, questions and answers.
And because it is aligned with the  
RecapVideo, extracted questions and answers can be grounded in 
specific segments of the movie and its corresponding dialogue.
We leverage this insight to introduce \movierecapsqa, a novel benchmark that features open-ended questions requiring multimodal reasoning over long-form, narrative video. Multiple levels of granularity of input video length are provided for each QA pair ranging from the full-length movie to just the RecapVideo segments relevant to the question.

Crucially, \movierecapsqa is designed to solve the core evaluation problems of open-ended questions in video. Specifically, we introduce an intermediary annotation layer of atomic facts---concise, verifiable statements---that are automatically derived from the RecapSummary and used to generate a QA pair. 
The facts, in turn, support our proposed a reference-free evaluation that uses an LLM judge grounded with question-specific facts to evaluate a generated answer.  The facts
provide a precise, text-based, verifiable representation of video content that enables the assessment of answer factuality, relevance, and coherence without relying on restrictive reference answers.

Using \movierecapsqa, we benchmark seven state-of-the-art
multimodal large language models (MLLMs), both proprietary and open-source, alongside human
participants. The dataset exposes several trends:
(i) semantic and reference-based metrics often rank models similarly with little separation,
whereas our reference-free
metric yields more meaningful separation and better alignment with human preferences;
(ii) vision-centric questions yield the lowest scores across all models;
(iii) proprietary model factuality \emph{improves} when visual input is removed, revealing a visual understanding gap;
and (iv) the primary bottleneck is visual perception, not reasoning, with fine-grained scene details showing the largest human--model gap.\footnote{All code is released, \href{https://sshaar.github.io/MovieRecapsQA/}{\movierecapsqa}.}

Our contributions can be summarized as:
\begin{enumerate}
    \item We introduce \textbf{\movierecapsqa}, a long-form,
    multimodal VideoQA benchmark comprising $\approx$8{,}200
    open-ended questions across 60 videos with
    modality-type labels (dialogue-centric, vision-centric,
    multimodal) and reasoning categories.

    \item We propose a \textbf{reference-free evaluation metric} for open-ended VideoQA that uses recap-derived atomic facts to ground an LLM judge in assessing factuality, relevance, and coherence without relying on a unique reference answer.

    \item We benchmark and analyze the performance of 7 multimodal LLMs and human annotators on \movierecapsqa, revealing a substantial human--model performance gap, a pronounced over-reliance on dialogue, and systematic failures on fine-grained visual perception.
\end{enumerate}

%% file: tables/table-compare.tex
\begin{table*}[htb!]
\centering
\small
\setlength{\tabcolsep}{6pt}
\footnotesize
\begin{tabular}{lccccc}
\toprule
\textbf{Dataset} & \textbf{Annotation} & \textbf{\# QAPairs} & \textbf{Avg. Len. (s)} & \textbf{Modality} & \textbf{Q Type}\\
\midrule
MovieQA~\cite{tapaswi_movieqa_2016} & Human & 6,462 & 203 & No & MC\\
TGIF-QA~\cite{jang_tgif-qa_2017} & Auto/Human & 165,165 & 3 & No & MC \& OE \\
TVQA~\cite{lei_tvqa_2019} & Human & 152,545 & 76 & No & MC \\
DramaQA~\cite{choi_dramaqa_2020} & Human & 17,983 & 91 & No & MC \\
MoVQA~\cite{zhang_movqa_2023} & Human & 21,953 & 992 & No & MC \\
CinePile~\cite{rawal_cinepile_2024} & Auto & 303,828 & 160 & No & MC \\
\rowcolor{gray!15}
\textbf{MovieRecapsQA (Ours)} & Auto & 8,231 & 660 & Yes & FF \\
\bottomrule
\end{tabular}
\caption{Comparison of existing video QA datasets. Our proposed dataset introduces multimodal distinctions and free-form question types, distinguishing it from prior benchmarks.
}
\end{table*}

%% file: sections/2.related-work.tex
\section{Related Work}
\label{sec:related_work}

\paragraph{Video Question Answering (VideoQA).} VideoQA was introduced as a proxy task to evaluate a model’s ability to understand and reason over video inputs. Existing VideoQA benchmarks typically adopt either a multiple-choice format (e.g., TVQA \cite{lei_tvqa_2019}, MovQA \cite{zhang_movqa_2023}, How2QA \cite{yang2022learninganswervisualquestions}, DramaQA \cite{choi_dramaqa_2020}) or an open-ended format (Video-Bench \cite{ning2023videobenchcomprehensivebenchmarktoolkit}, MVBench \cite{li2024mvbenchcomprehensivemultimodalvideo}, EgoSchema \cite{mangalam2023egoschemadiagnosticbenchmarklongform}, CinePile \cite{rawal_cinepile_2024}). Open-ended benchmarks, however, remain both more challenging and far less common, as evaluation for free-form responses is still an open problem \cite{ataallah2025infinibenchbenchmarklargemultimodal}.

Moreover, there has been growing interest in multimodal VideoQA \cite{zhang_movqa_2023, rawal_cinepile_2024}, where models receive both video and textual context -- and questions typically require integrating information across visual and textual modalities to generate accurate responses. However, curating such datasets is far more expensive, and they often need to be partially or fully automatically generated, with some generation methods limiting the resulting QA pairs to either visual-only or dialogue-only questions.

\paragraph{Open-Ended Question Answering Evaluation}
Evaluating the quality of open-ended, free-form text remains one of the most significant challenges in QA, and other generation tasks.
For years, the field relied on n-gram overlap metrics, such as ROUGE\cite{lin_rouge_2004}, BLEU\cite{papineni-etal-2002-bleu}, METEOR\cite{banerjee-lavie-2005-meteor}. 
Later embedding-based metrics such as BERTScore \cite{zhang_bertscore_2020} and BARTScore were introduced, though they showed limited alignment with human judgments \cite{fabbri2021summevalreevaluatingsummarizationevaluation}.
More recently, open-ended QA benchmarks have exclusively used LLM-as-a-judge metrics (e.g., HELMET \cite{yen2025helmet} and GEval \cite{liu_g-eval_2023}) to evaluate the relevance and coherence of model answers. 
Both approaches, however, require reference answers.
To our knowledge, there are no benchmarks that evaluate model answers without relying on a reference, a limitation that prevents the creation of larger and more cost-effective datasets.

In other QA settings (e.g., text-based QA), the factuality of an answer is a critical evaluation dimension but remains difficult to measure reliably \cite{kryscinski-etal-2020-evaluating, fabbri-etal-2022-qafacteval}. 
Early approaches relied on NLI models or entity matching \cite{goyal-durrett-2020-evaluating}, yet these methods consistently failed to capture factuality with strong human alignment \cite{kamoi-etal-2023-shortcomings}. 
More recently, LLM-as-judge factuality metrics (i.e., FactScore\cite{factscore} and VeriScore\cite{song_veriscore_2024}) assess factuality without requiring a reference answer more reliably. 
Since factuality concerns the truthfulness of the answer—not its similarity to a reference—these methods instead verify claims against the input text context, with some work suggesting that context quality affects scoring\cite{ramprasad2025automaticfactualitymetricsmeasure}.
This is particularly difficult to achieve in VideoQA, as using raw video as the verification context is unreliable (with MLLM) and computationally expensive. 
To our knowledge, no VideoQA dataset explicitly measures factuality.

%% file: sections/3.dataset.tex
\section{\movierecapsqa Benchmark}
\label{sec:dataset}

\begin{figure*}[htb!]
    \centering
    \includegraphics[width=\linewidth]{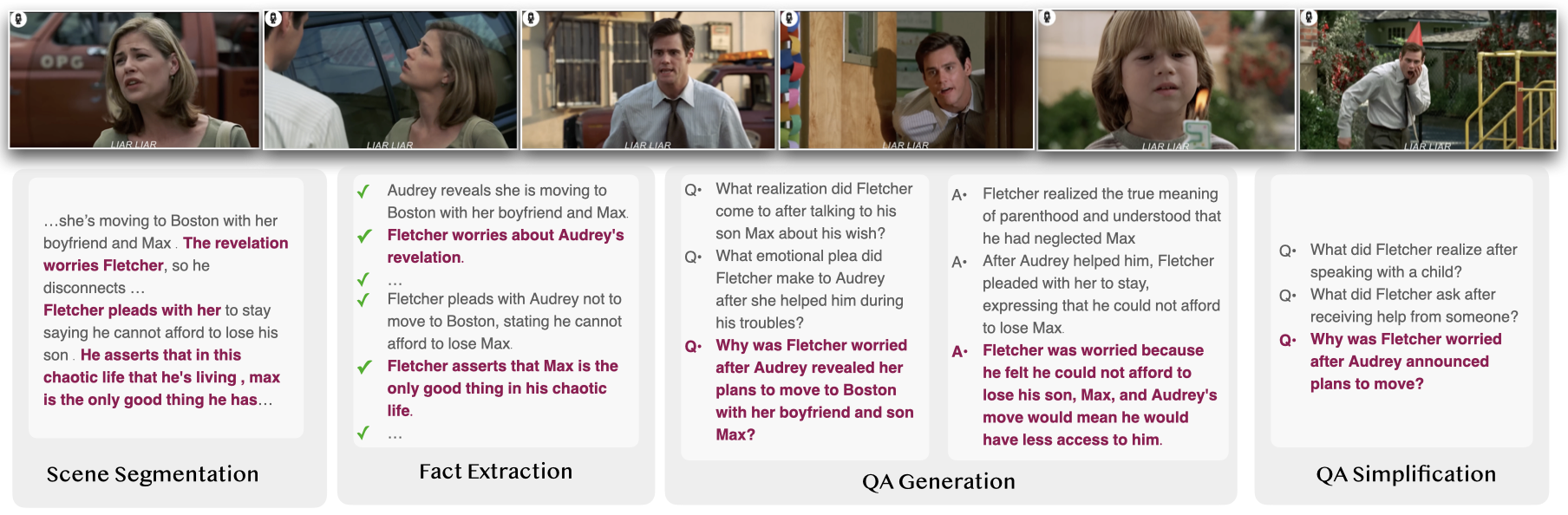}
    \caption{\textbf{QA Generation Pipeline.} Example question-answer pairs from recap video 6Tfmy3uGTmQ (for \textit{Liar Liar}) on the recap segment 
    ``00:06:50--00:12:42" (and ``00:31:44--00:50:12" from the movie).
    The red-highlighted text indicates the recap-segment input used to extract facts and generate the corresponding QA pair.}
    \label{figure:dataset_qa_generation}
\end{figure*}

We construct our dataset automatically by leveraging a widely available and increasingly popular genre of YouTube content known as recap videos. 
These videos narrate the full storyline of a movie—typically in an 8–15 minute continuous format—while replaying key scenes. 
Unlike Wikipedia Synopsis or IMDb plot summaries, which are often high-level and omit substantial narrative detail, recap videos provide dense, scene-by-scene coverage of the film’s major events, characters, and plot developments.

Additionally, unlike other VideoQA datasets built from movies (e.g., MovieQA, TVQA, MovQA), the recap videos directly pair each narrated event with the corresponding movie shot(s) through visual-only representation (i.e., no audio). 
This tight coupling between narration and visuals allows for accurate summary–movie alignment and enables a more accurate pairing between the question, the movie dialogue, and the question-specific facts derived from the RecapSummary. The dataset was constructed in two main steps: (1) Collection \& Alignment, and (2) QA Generation.\footnote{We address all copyright issues in Appendix~\ref{sec:appendix_copyright}.}

\input{tables/dataset-statistics} 

\subsection{\movierecapsqa Collection \& Alignment}

We begin by selecting a set of 60 films released between 1980 and 2024, spanning both widely known titles (e.g., Avatar) and more niche works (e.g., Year One).\footnote{We intentionally include movies with publicly available scripts so that future work can incorporate script-based annotations and build further extensions on top of our benchmark.}
To source recap content, we compile a list of the top 10 most popular YouTube recap channels, which we treat as trusted providers of high-quality summaries. Using the YouTube API, we retrieve candidate RecapVideos for each movie and collect the top five search results originating from these trusted channels. Finally, we manually filter these videos, verifying that each recap is indeed aligned with the corresponding movie’s storyline and title.

Next produce a scene-level alignment of each RecapVideo to its corresponding movie.
We first apply SceneDetect \citep{islam2023efficientmoviescenedetection} to segment both the full movie and its corresponding RecapVideo into scenes. We group all consecutive frames belonging to the same scene.
Then we embed the first and last three seconds of footage for every detected scene in both the movie and the RecapVideo, and compute cosine similarity to match corresponding shots. We use SlowFast \cite{feichtenhofer2019slowfastnetworksvideorecognition} for the embeddings. 
Because some RecapVideos are not strictly chronological, we additionally perform a lightweight statistical alignment step to enforce a semi-chronological ordering of matched scenes. Further alignment details and results can be found in  Appendix \ref{sec:appendix_dataset_alignment}. 

Using this process, we obtain an alignment of the RecapVideo not only with the movie video but also with the corresponding dialogue via the available movie subtitles. Because the RecapVideo is itself aligned with the RecapSummary, we also obtain an alignment of the summary with the movie and its dialogue.

\subsection{\movierecapsqa QA Generation}

To construct the question–answer pairs, we divide each RecapSummary into segments and use prompting to extract all facts from each recap segment via LLM-based prompting. We then instruct an LLM to generate QA pairs that rely on one or more facts within a single segment. However, because these extracted facts tend to be highly verbose, the resulting QA pairs are often too easy: they reveal excessive information about the scene. In addition, questions frequently include character names, which can hinder alignment with subtitles (since dialogue excerpts may not mention the character’s name at the corresponding moment). To address these issues, we additionaly generated simplified QA pairs using a dedicated LLM prompt designed to abstract away identifying details. All dataset construction steps are performed using \textsc{GPT-4.1}. All prompts regarding the QA generation can be found in Appendix~\ref{sec:appendix_dataset_creation_details}.

Figure~\ref{figure:dataset_qa_generation} illustrates QA generation for Segment~4 of the Liar Liar RecapVideo: fact extraction produces 32 facts (only four are shown); then verbose QA pairs and their simplified counterparts are generated.  We retain the answer from the verbose QA pair and the question from the simplified QA pair in our final QA set.


\subsection{Statistics}
The \movierecapsqa dataset is built on 60 films, resulting in a corpus of 8,231 open-ended question-answer pairs.  Each question is aligned with with a specific segment from a movie recap video, the corresponding (video-only) segment from the original full-length movie, and the corresponding movie subtitles.  A summary of the dataset's key statistics is presented in Table~\ref{tab:question_stats}.

Questions are categorized by the primary modality required to answer them and the type of reasoning involved.  The distribution of question modalitites is diverse, as shown in Table~\ref{tab:question_stats}, which forces the models to use different types of reasoning.  The breakdown of different question categories, based on the CinePile taxonomy, is also detailed in Table~\ref{tab:question_stats}. All prompts used to categories the dataset can be found in Appendix~\ref{sec:appendix_question_cats_modality}.

\input{tables/modalities_and_types}

%% file: tables/dataset-statistics.tex
\definecolor{tablegray}{gray}{0.93}   
\definecolor{rowgray}{gray}{0.97}     

\definecolor{tablegray}{gray}{0.93}

\sisetup{
  group-separator={,},      
  group-minimum-digits=4,   
  detect-weight=true,
  detect-family=true,
}

\begin{table}[thb!]
\centering
\small
\setlength{\tabcolsep}{6pt}
\renewcommand{\arraystretch}{1.1}
\rowcolors{3}{rowgray}{white} 
\resizebox{.9\columnwidth}{!}{
\begin{tabularx}{\columnwidth}{@{}>{\raggedright\arraybackslash}X r@{}}
\toprule
\rowcolor{tablegray}
\textbf{Statistic} & \textbf{Value} \\
\midrule
\rowcolor{white}
\multicolumn{2}{@{}l}{\textbf{Dataset Size}} \\
Total Films \& Recap Videos         & 60 \\
\textbf{Total QA Pairs}             & \textbf{8{,}231} \\
\textbf{Total Aligned Atomic Facts} & \textbf{16{,}462} \\
\midrule
\rowcolor{white}
\multicolumn{2}{@{}l}{\textbf{Temporal Statistics}} \\
Avg. Length of Full Recap Video     & $\sim$660 s (11 min) \\
Avg. Length of Full Movie           & $\sim$6{,}446 s (107 min) \\
Avg. Segment Length (Recap)         & $\sim$73 s \\
Avg. Aligned Segment (Full)         & $\sim$863 s (14 min) \\
\midrule
\rowcolor{white}
\multicolumn{2}{@{}l}{\textbf{Facts and Questions}} \\
Avg. Total Facts from Segments      & $\sim$12 \\
Avg. Facts per Video                & $\sim$222 \\
Avg. Questions from Segments        & $\sim$172 \\
\bottomrule
\end{tabularx}
}
\caption{Overall statistics for the \movierecapsqa dataset.}
\label{tab:dataset_stats}
\end{table}

%% file: tables/modalities_and_types.tex
\begin{table}[thb!]
    \centering
    \small
    \begin{tabular}{l r}
        \toprule
        \rowcolor{gray!12}
        \multicolumn{2}{l}{\textbf{Question Modality}} \\
        Dialogue-centric & 2{,}932 \\
        Multimodal       & 3{,}525 \\
        Vision-centric   & 1{,}774 \\
        \midrule
        \rowcolor{gray!12}
        \multicolumn{2}{l}{\textbf{Question Categories}} \\
        Narrative and Plot Analysis (NPA)              & 3{,}294 \\
        Character and Relationship Dynamics (CRD)      & 3{,}149 \\
        Thematic Exploration (TH)                      & 677 \\
        Setting and Technical Analysis (STA)         & 655 \\
        Temporal Reasoning (TEMP)                      & 456 \\
        \midrule
        \rowcolor{gray!18}
        \textbf{Total} & \textbf{8{,}231} \\
        \bottomrule
    \end{tabular}
    \caption{Question statistics for \movierecapsqa: modality distribution (top) and categories (bottom) over 8{,}231 total questions.}
    \label{tab:question_stats}
\end{table}

%% file: sections/4.evaluation-metrics.tex
\section{Evaluation Metrics}
\label{sec:evaluation-metrics}



Evaluating free-form open-ended VideoQA is fundamentally harder than evaluating multiple-choice or short-answer formats. A correct open-ended answer must satisfy multiple criteria simultaneously: it must be factually accurate with respect to the video content, relevant to the question being asked, and internally coherent. No single existing metric captures all three dimensions, and most fail to capture even one reliably. We first survey the shortcomings of standard metrics, then introduce our reference-free evaluation framework designed to address them.



\subsection{Baseline Metrics}

\paragraph{Semantic Evaluation Metrics}

We first establish baselines using standard semantic similarity metrics that compare model-generated responses against reference answers at the lexical and embedding level.
\textbf{ROUGE} \cite{lin_rouge_2004} measures n-gram overlap; \textbf{BERTScore} \cite{zhang_bertscore_2020} computes embedding-based similarity; and \textbf{BARTScore} \cite{yuan_bartscore_2021} uses a seq2seq model to score conditional likelihood. These metrics capture surface form and paraphrastic similarity but not factual correctness. 

\paragraph{Reference-Based Evaluation Metrics}
To assess factual correctness in long-form answers we employ reference-based LLM judges that leverage the reasoning capabilities of LLMs to evaluate response quality.

We include two reference-based LLM judges: \textbf{G-Eval} \cite{liu_g-eval_2023},  a framework that rates the coherence and consistency of an answer given a reference, and \textbf{HELMET}\cite{yen2025helmet}, which targets two dimensions --- \emph{Fluency}, which captures grammatical correctness and coherence, and \emph{Correctness}, which is designed to measure factual agreement with the reference answer on a 0--3 scale.  HELMET Correctness serves as our primary reference-based baseline as it explicitly targets factual accuracy.  However, it still relies on a single gold standard answer and cannot verify whether a model's response is grounded in the underlying video content.

\subsection{Reference-Free Evaluation Metric}
\label{subsec:fact-metric}

The limitations above motivate our central evaluation contribution: a reference-free LLM judge that evaluates whether a response is factually grounded in the video content, independent of how any particular reference answer is formulated.


We first leverage our recap pipeline to construct a \emph{textual layer of atomic facts} that serves as a compact, verifiable representation of the video content.

For each question $q$, we collect a set of \emph{atomic facts}
$\mathcal{F}_q = \{f_1, \dots, f_K\}$ derived from the aligned recap
summary for the corresponding video segment.
Each atomic fact is a short, standalone proposition about the movie
(e.g., ``\textit{Tyler threatens to destroy the narrator's apartment}''),
written so that its truth can be directly verified from the recap.
In addition to the atomic facts extracted for each questions, we also extract a set of \emph{claims} for each model response $\mathcal{C}_r = \{c_1, \dots, c_K\}$.
Given a question $q$, the model claims ${C}_r$, the associated atomic
facts $\mathcal{F}_q$, and the subtitles for the segment our LLM judge is prompted to evaluate $a$ along three dimensions:

\begin{itemize}
    \item \textbf{Factuality} ($s_{\mathrm{fact}}$): to what extent the
    claims in $a$ are supported (or contradicted) by $\mathcal{F}_q$.
    \item \textbf{Relevance} ($s_{\mathrm{rel}}$): whether $a$ directly
    addresses $q$ and avoids introducing unsupported, off-topic content.
\end{itemize}

All reference-free evaluations are performed using GPT-4.1 mini as the LLM judge. We select this model for its cost efficiency given the scale of evaluation, as scoring ~8.2K questions across seven models and three dimensions requires a substantial number of API calls. We verified that GPT-4.1 mini produces scores comparable to GPT-4.1 on our evaluation dimensions.

The judge returns integer scores in the range $0$--$5$ for each dimension,
where higher scores indicate better performance. 
Full prompts can be found in Appendix~\ref{sec:appendix_evalaution_prompts}.

We exclude the coherence metric from the main results, as the answers are too short to meaningfully exhibit internal contradictions.
Results on the coherence dimension are reported in Appendix~\ref{sec:appendix_coherence}.

%% file: sections/6.experiment.tex
\input{tables/main_results}

\section{Experiments}
\label{sec:setup}

Our goal is to evaluate how current multimodal LLMs handle
long-form, open-ended VideoQA under controlled, comparable
conditions. This section describes how we construct model
inputs, how we prompt each system, and how we run our
ablations.

\subsection{Models}
We evaluate a diverse set of multimodal models, spanning both proprietary and open-source systems. Our selection includes leading proprietary models: GPT-4o\cite{openai_gpt-4o_2024}, Gemini 2.5 Flash\cite{google2025gemini25flash}, Claude 3.5 Sonnet\cite{anthropic2024claude35sonnet}, and Amazon Nova Lite\cite{Intelligence2024}, which represent the current frontier in commercial video understanding capabilities. We complement these with open-source alternatives: LLaVA NeXT-Video\cite{li_llava-next_2024}, MiniCPM-o\cite{hu_minicpm_2024}, and Qwen 2.5-VL\cite{bai_qwen25-vl_2025}. 
All models are evaluated in a zero-shot
setting with a standardized prompting scheme and input
format.

\subsection{Input Formats and Modalities}

Each question in \movierecapsqa is aligned with a specific
recap segment, its corresponding full-movie segment, and the
matched dialogue (subtitles + script) for that interval
(Section~\ref{sec:dataset}). Unless otherwise noted, all
models are evaluated using the recap segment as the visual
source.\footnote{The dataset design enables multiple task formulations, with variable video length (clip, recap, full movie) and variable text length (segment-level or full dialogue). Further details can be found in Appendix \ref{sec:appendix_alternate}.}

\vspace*{-.2\baselineskip}
\paragraph{Visual input.}
For models that accept raw video files (e.g., Gemini 2.5
Flash, Amazon Nova Lite), we pass the aligned recap segment
as a short MP4 clip. For frame-based models (e.g., GPT-4o,
Claude 3.5 Sonnet, LLaVA-NeXT-Video, MiniCPM-o,
Qwen2.5-VL), we uniformly sample frames from the recap
segment and pass them as an ordered image sequence. 
We use a different number of frames for each model, up to the maximum context length allowed by that model. 

\vspace*{-1\baselineskip}
\paragraph{Dialogue input.}
We use the aligned subtitles from the same temporal window as the visual input.

\vspace*{-0.9\baselineskip}
\paragraph{Multimodal, frames-only, and dialogue-only
conditions.}
By default, models in the \emph{multimodal} setting receive
both the visual input (video or frames) and the aligned
dialogue. For our ablations, we also evaluate two restricted
conditions:
(i) \emph{frames-only}, where the model receives only the
visual input and no text, and
(ii) \emph{dialogue-only}, where the model receives only the
subtitle snippet without any images or video.
These three conditions allow us to tease apart how much each
system relies on vision versus language, and whether
multimodal input improves or harms factual accuracy.

\subsection{Human Study}
\label{sec:human-study}
To establish a human upper bound and to validate our evaluation metrics, we conducted a human study.  We randomly sampled 118 questions from our dataset, spanning all three modalities (dialogue-centric, vision-centric, multimodal)  These questions were presented to five human participants, who provided open-ended answers under the same conditions as models:  each participant saw the question, the recap segment, and the aligned subtitles for that segment. 

We then scored human answers using HELMET Correctness and our fact-based Factuality and Relevance metrics. For each question, we compute both the \emph{average} human score (averaged across participants) and the \emph{best}
human score (maximum across participants).

%% file: tables/main_results.tex
\begin{table*}[thb!]
\centering
\small
\setlength{\tabcolsep}{4pt}
\resizebox{\textwidth}{!}{
\begin{tabular}{lcccccccc}
\toprule
 & \multicolumn{3}{c}{\textbf{Semantic Metrics}} 
 & \multicolumn{3}{c}{\textbf{Reference-Based Evaluation}} 
 & \multicolumn{2}{c}{\textbf{Reference-Free Evaluation}} \\
\cmidrule(lr){2-4}\cmidrule(lr){5-7}\cmidrule(lr){8-9}
& & & & & \textbf{HELMET} & \textbf{HELMET} &  &  \\
\textbf{Model} & \textbf{ROUGE-L} & \textbf{BERTScore} & \textbf{BARTScore} &
\textbf{G-Eval} & \textbf{Fluency} & \textbf{Correctness} &
\textbf{Factuality} & \textbf{Relevance} \\
\midrule
LLaVA-NeXT-Video  
& $0.23_{\pm 0.01}$ 
& $0.65_{\pm 0.01}$ 
& $0.03_{\pm 0.00}$ 
& $0.26_{\pm 0.03}$ 
& $0.96_{\pm 0.04}$ 
& $0.98_{\pm 0.89}$ 
& $2.96_{\pm 1.97}$ 
& $3.35_{\pm 1.47}$ \\

Mini-CPM-o  
& $0.24_{\pm 0.02}$ 
& $0.65_{\pm 0.01}$ 
& $0.04_{\pm 0.00}$ 
& $0.30_{\pm 0.04}$ 
& $0.94_{\pm 0.05}$ 
& $1.27_{\pm 1.08}$ 
& $3.21_{\pm 2.04}$ 
& $3.61_{\pm 1.47}$ \\

Qwen2.5VL  
& $0.26_{\pm 0.02}$ 
& $0.67_{\pm 0.01}$ 
& $0.04_{\pm 0.00}$ 
& $0.31_{\pm 0.03}$ 
& $0.97_{\pm 0.03}$ 
& $1.23_{\pm 0.98}$ 
& $3.47_{\pm 1.98}$ 
& $3.83_{\pm 1.41}$ \\

Amazon Nova Lite  
& $0.28_{\pm 0.02}$  
& $0.69_{\pm 0.01}$  
& $0.05_{\pm 0.00}$  
& $0.32_{\pm 0.04}$ 
& {$0.99_{\pm 0.01}$} 
& $1.29_{\pm 1.03}$ 
& $3.53_{\pm 1.96}$ 
& $3.93_{\pm 1.35}$ \\

Claude 3.5 Sonnet  
& $0.22_{\pm 0.02}$ 
& $0.63_{\pm 0.01}$ 
& $0.05_{\pm 0.00}$ 
& $0.37_{\pm 0.05}$ 
& $0.98_{\pm 0.02}$ 
& $1.35_{\pm 1.42}$ 
& $3.76_{\pm 1.80}$ 
& $3.92_{\pm 1.19}$ \\

Gemini-2.5-Flash  
& $0.22_{\pm 0.02}$ 
& $0.63_{\pm 0.01}$ 
& $0.05_{\pm 0.00}$ 
& {$0.38_{\pm 0.05}$} 
& $0.95_{\pm 0.05}$ 
& {$1.82_{\pm 1.33}$} 
& $3.26_{\pm 2.35}$ 
& $3.70_{\pm 1.57}$ \\

GPT-4o  
& $0.28_{\pm 0.03}$ 
& $0.68_{\pm 0.01}$ 
& $0.05_{\pm 0.01}$ 
& $0.37_{\pm 0.06}$ 
& $0.94_{\pm 0.05}$ 
& $1.43_{\pm 1.18}$ 
& $3.99_{\pm 2.01}$ 
& $3.97_{\pm 1.92}$ \\
\midrule

\rowcolor{blue!10}
Avg. Human*  
& $0.16_{\pm 0.01}$ 
& $0.88_{\pm 0.00}$ 
& -- 
& -- 
& $0.94_{\pm 0.06}$ 
& $0.98_{\pm 1.06}$ 
& $4.01_{\pm 1.70}$ 
& $4.01_{\pm 1.34}$ \\

\rowcolor{blue!10}
Best Human* 
& $0.19_{\pm 0.01}$ 
& $0.87_{\pm 0.00}$ 
& -- 
& -- 
& $0.93_{\pm 0.06}$ 
& $1.26_{\pm 1.21}$ 
& {$4.59_{\pm 0.63}$} 
& {$4.53_{\pm 0.76}$} \\
\bottomrule
\end{tabular}
}
\caption{
\textbf{Models Performance.} This table reports model performance on our benchmark across semantic, reference-based, and reference-free metrics. We additionally include \textsc{Human*} performance on a sampled set of 118 questions. For each metric, we report the mean score across all benchmark questions $\pm$ variance. HELMET scores range from 0–3, our reference-free metrics from 0–5, and all remaining metrics are normalized to 1.
}
\label{tab:metrics}
\end{table*}

%% file: sections/7.analysis.tex
\section{Results \& Discussion}
\label{sec:experiments-analysis}

This section presents an analysis of model performance on the \movierecapsqa benchmark. 
Table~\ref{tab:metrics} reports overall performance across all metrics and models, and 
Table~\ref{tab:model_relevance_types_categories_mean} presents ablation results across 
question categories and reasoning types; further ablation details are provided in 
Appendix~\ref{sec:appendix_detailed_ablation_results}.

\subsection{The Metric Divergence Problem}
\vspace*{-.2\baselineskip}
\paragraph{Observation \#1: Semantic scores fail to discriminate between models.}
Semantic metrics show virtually no discrimination across models; for example, \textsc{ROUGE-L}
ranged only $0.22$--$0.28$, \textsc{BERTScore} $0.63$--$0.69$, and \textsc{BARTScore}
$0.03$--$0.05$, as shown in Table~\ref{tab:metrics}. Moreover, the per-model variance of each
metric is $\leq 0.03$, indicating that these metrics assign nearly identical scores to every
answer produced by the same model, regardless of question or answer content. Together, this
suggests that n-gram overlap and embedding-based similarity fail to capture anything beyond
surface-level token matching.

\vspace*{-1\baselineskip}
\paragraph{Observation \#2: Reference-based LLM judges do not align with human preference.}
While reference-based metrics yield better model separation than semantic metrics, they do not
align with human preference, Table~\ref{tab:metrics}. For example, \textsc{HELMET-Correctness}
rates \textsc{MiniCPM-o}, $1.27$, above the best, $1.26$, and average human, $0.98$. Similarly \textsc{LLaVA-NeXT-Video}, $0.98$, is scored on par with the average human. This suggests that reference-based judges inflate
model scores and fail to reflect the true gap between model and human performance.

\vspace*{-1.2\baselineskip}
\paragraph{Observation \#3: Reference-free metric produces human-aligned model separation.}
In contrast to both prior metric families, our reference-free metrics produce a wider and more
meaningful spread, Table~\ref{tab:metrics}. \textsc{Factuality} scores range from
$2.96$--$3.99$ across models, complimented with $4.01$ and $4.59$ for average and best human. 
Furthermore, the higher per-model variance
of our metric ($1.97$--$2.35$ on \textsc{Factuality}) relative to reference-based metrics
($0.89$--$1.42$) confirms that our scores reflect genuine differences in answer quality across
questions rather than model-level tendencies.

\subsection{The Modality Performance Gap}
\label{subsec:perf_modality}

\vspace*{-.2\baselineskip}
\paragraph{Observation \#4: Vision-centric questions yield the lowest Factuality scores across all models.}
Questions requiring visual understanding, whether \textit{vision-centric} or \textit{multimodal}, score lowest on \textsc{Factuality} for both models and
humans, Table~\ref{tab:model_relevance_types_categories_mean}. 
Proprietary models drop from
$3.63$ (dialogue) to $3.15$ (vision); open-source models from $3.21$ to $3.05$. Humans follow
the same trend, falling from $4.17$ to $3.84$. 
This confirms that \movierecapsqa poses a genuine
visual challenge: getting the facts right is hard not because models are weak, but because
visual information is difficult to extract even for humans watching the same video.

\vspace*{-1\baselineskip}
\paragraph{Observation \#5: Models stay on-topic but fail to get the facts right.}
Unlike Factuality, Relevance scores remain stable across modalities for proprietary, $3.84$--$3.63$, and open-source, $3.52$--$3.61$, 
indicating that models consistently attend to the correct scene regardless of modality. 
Yet humans score substantially higher on Relevance across modality (${\approx}4.0+$), revealing a clear dissociation: models discuss the right content but cannot extract the precise
visual facts needed to answer correctly, i.e., they know \textit{where} to look, but not
\textit{what} to say.

\input{tables/ablation_small}

\subsection{The Visual Information Gap}
\label{subsec:visual_gap}

\vspace*{-.2\baselineskip}
\paragraph{Observation \#6: Removing visual input improves proprietary model performance.}
Proprietary models score higher under dialogue-only than full-modality input across all question
types and both metrics (Table~\ref{tab:model_relevance_types_categories_mean}). \textsc{Factuality}
improves by $+0.49$, $+0.73$, and $+0.70$ on dialogue, vision-centric, and multimodal questions
respectively, with \textsc{Relevance} following the same trend. The gains are largest on
vision-centric questions, questions relying on visual cues only, indicating that
these models do not extract useful signal from video frames. 
Rather than helping, visual input actively disrupts their predictions, pulling them away from the textual priors they would otherwise rely on.

\vspace*{-1\baselineskip}
\paragraph{Observation \#7: Proprietary models lose their advantage when given only frames.}
Under full-modality input, proprietary models lead open-source across all question types, but this advantage collapses under frames-only, Table~\ref{tab:model_relevance_types_categories_mean}. 
On vision-centric questions, open-source models score better than proprietary ($3.15$ vs.\ $3.06$). 
We also observe that open-source models show smaller performance drops under frames-only in comparison to proprietary models.
Combined with Observation~\#6, this suggests that proprietary superiority is driven by stronger language priors, not better visual understanding.

\subsection{Performance Across Question Categories}

\vspace*{-.1\baselineskip}
\paragraph{Observation \#8: The primary performance bottleneck is visual perception, not reasoning type.}
Performance across question categories follows a clear pattern: categories that reward dialogue comprehension (CRD, NPA) are easiest for models ($3.53$, $3.41$ proprietary Factuality), while STA (i.e., requiring fine-grained perception of lighting, camera work, and environmental detail) is hardest by a wide margin ($2.98$ proprietary, $2.83$ open-source). 
Interestingly, humans show no such drop ($4.15$), making the human--model gap on STA ($1.17$--$1.32$) roughly double that of CRD or NPA. 
The modality ablations sharpen the point: on STA, proprietary models given dialogue-only \emph{improve} by $+0.63$ Factuality, while those given frames-only \emph{decline} by $-0.19$, indicating that visual encoders actively hurt on questions demanding the most from them.

%% file: tables/ablation_small.tex
\begin{table*}[thb!]
\centering
\small
\setlength{\tabcolsep}{4pt}
\begin{tabular}{l | ccc | ccccc}
\toprule
\multicolumn{1}{c}{ } & \multicolumn{3}{c}{\textbf{Question Types}} & \multicolumn{5}{c}{\textbf{Question Categories}} \\
\cmidrule(lr){2-4} \cmidrule(lr){5-9}
\multicolumn{1}{c}{\textbf{Model}}  & \textbf{Dialogue} & \textbf{Scene} & \textbf{Multimodal} & \textbf{CRD} & \textbf{NPA} & \textbf{STA} & \textbf{TEMP} & \textbf{TH} \\
\midrule
\multicolumn{9}{c}{\textbf{Relevance Score}} \\
\midrule
\rowcolor{blue!5}
\textbf{Open-Source Models} & \ \ \ 3.61 & \ \ \ 3.53 & \ \ \ 3.52 & \ \ \ 3.53 & \ \ \ 3.52 & \ \ \ 3.58 & \ \ \ 3.70 & \ \ \ 3.72 \\
(only frames) &  {\color{BetterGreen}$+0.06$} & {\color{BetterGreen}$+0.14$} & {\color{BetterGreen}$+0.15$} & {\color{BetterGreen}$+0.15$} & {\color{BetterGreen}$+0.07$} & {\color{BetterGreen}$+0.08$} & {\color{BetterGreen}$+0.00$} & {\color{BetterGreen}$+0.11$} \\
(only dialogue) &  {\color{BetterGreen}$+0.05$} & {\color{BetterRed}$-0.09$} & {\color{BetterGreen}$+0.08$} & {\color{BetterGreen}$+0.09$} & {\color{BetterGreen}$+0.03$} & {\color{BetterRed}$-0.14$} & {\color{BetterRed}$-0.16$} & {\color{BetterGreen}$+0.01$} \\
\rowcolor{blue!5}
\textbf{Proprietary Models} & \ \ \ 3.84 & \ \ \ 3.63 & \ \ \ 3.83 & \ \ \ 3.82 & \ \ \ 3.74 & \ \ \ 3.53 & \ \ \ 3.73 & \ \ \ 3.86 \\
(only frames) &  {\color{BetterRed}$-0.18$} & {\color{BetterGreen}$+0.01$} & {\color{BetterRed}$-0.11$} & {\color{BetterRed}$-0.14$} & {\color{BetterRed}$-0.09$} & {\color{BetterGreen}$+0.11$} & {\color{BetterRed}$-0.11$} & {\color{BetterRed}$-0.17$} \\
(only dialogue) &  {\color{BetterGreen}$+0.37$} & {\color{BetterGreen}$+0.24$} & {\color{BetterGreen}$+0.30$} & {\color{BetterGreen}$+0.33$} & {\color{BetterGreen}$+0.32$} & {\color{BetterGreen}$+0.31$} & {\color{BetterGreen}$+0.15$} & {\color{BetterGreen}$+0.35$} \\
\rowcolor{blue!5}
\textbf{Human*} & \ \ \ 4.27 & \ \ \ 3.97 & \ \ \ 4.00 & \ \ \ 4.05 & \ \ \ 3.98 & \ \ \ 4.41 & -- & \ \ \ 4.11 \\
\midrule
\multicolumn{9}{c}{\textbf{Factuality Score}} \\
\midrule
\rowcolor{blue!5}
\textbf{Open-Source Models} & \ \ \ 3.21 & \ \ \ 3.05 & \ \ \ 3.11 & \ \ \ 3.19 & \ \ \ 3.13 & \ \ \ 2.83 & \ \ \ 3.15 & \ \ \ 3.02 \\
(only frames) &  {\color{BetterRed}$-0.14$} & {\color{BetterGreen}$+0.08$} & {\color{BetterRed}$-0.02$} & {\color{BetterRed}$-0.07$} & {\color{BetterRed}$-0.03$} & {\color{BetterGreen}$+0.12$} & {\color{BetterRed}$-0.06$} & {\color{BetterRed}$-0.09$} \\
(only dialogue) &  {\color{BetterRed}$-0.01$} & {\color{BetterGreen}$+0.02$} & {\color{BetterGreen}$+0.10$} & {\color{BetterGreen}$+0.07$} & {\color{BetterGreen}$+0.08$} & {\color{BetterRed}$-0.05$} & {\color{BetterRed}$-0.16$} & {\color{BetterGreen}$+0.03$} \\
\rowcolor{blue!5}
\textbf{Proprietary Models} & \ \ \ 3.63 & \ \ \ 3.15 & \ \ \ 3.46 & \ \ \ 3.53 & \ \ \ 3.41 & \ \ \ 2.98 & \ \ \ 3.23 & \ \ \ 3.38 \\
(only frames) &  {\color{BetterRed}$-0.48$} & {\color{BetterRed}$-0.09$} & {\color{BetterRed}$-0.25$} & {\color{BetterRed}$-0.29$} & {\color{BetterRed}$-0.22$} & {\color{BetterRed}$-0.19$} & {\color{BetterRed}$-0.26$} & {\color{BetterRed}$-0.38$} \\
(only dialogue) &  {\color{BetterGreen}$+0.49$} & {\color{BetterGreen}$+0.73$} & {\color{BetterGreen}$+0.70$} & {\color{BetterGreen}$+0.64$} & {\color{BetterGreen}$+0.69$} & {\color{BetterGreen}$+0.63$} & {\color{BetterGreen}$+0.61$} & {\color{BetterGreen}$+0.59$} \\
\rowcolor{blue!5}
\textbf{Human*} & \ \ \ 4.17 & \ \ \ 3.84 & \ \ \ 3.98 & \ \ \ 4.07 & \ \ \ 3.86 & \ \ \ 4.15 & -- & \ \ \ 4.14 \\

\bottomrule
\end{tabular}
\caption{\textbf{Ablation Experiments.} We report the average performance of an open-source model and a proprietary model on our proposed reference-free metric, broken down by question types and categories. We include ablations where models are prompted with only video frames, only dialogue, or the full context (blue rows). We also report the average performance of \textit{Humans*} on each question type and category using the sampled set of 118 questions.}
\label{tab:model_relevance_types_categories_mean}
\end{table*}

%% file: sections/8.conclusion.tex
\section{Conclusion}
\label{sec:conclusion}
We proposed \movierecapsqa, a multimodal open-ended VideoQA benchmark built from aligned recap videos, full-length movies, subtitles, and summaries. By leveraging recap videos as an intermediate representation, our dataset supports questions that require reasoning over both visual and linguistic context while preserving the narrative structure of the underlying film. In addition, we proposed on atomic fact-based, reference-free evaluation framework that scores answers on factuality, coherence, and relevance without relying on a single reference response.

Taken together, our analyses highlight three broader implications for long-form VideoQA:
\begin{enumerate}
    \item {\textbf{Metrics matter.} Conventional semantic metrics and reference-based LLM judges are not reliable indicators of factual correctness in open-ended VideoQA. Our fact-based metric better separates models and humans and exposes genuine weaknesses, suggesting that future benchmarks should incorporate similar fact-grounded evaluation rather than relying solely on reference answers.}
	\item{\textbf{Multimodality is still brittle.} The visual information gap—where removing frames improves factuality—indicates that current MLLMs do not robustly integrate video and dialogue. Progress on visual encoders alone is not enough; we need architectures and training strategies that force models to ground language in visual evidence rather than treating subtitles as the primary source of truth.}
	\item{\textbf{Recap videos are a powerful but underused signal.} By aligning recap summaries, recap videos, and full movies, \movierecapsqa provides a scalable source of atomic facts and multimodal context that can drive both better models and better evaluation (through fact-based metrics).}
\end{enumerate}
We hope these findings encourage future work on (i) architectures that explicitly ground answers in both videos and dialogue, (ii) training objectives that reward visual-textual consistency rather than mere fluency, and (iii) new benchmarks that, like \movierecapsqa, pair long-form narrative videos with auxiliary textual signals rich enough to support atomic fact extraction and reference-free evaluation.

%% file: supplementary/A.copyright.tex
\section{Copyright}
\label{sec:appendix_copyright}

To ensure that \movierecapsqa is released in full compliance with copyright law and platform terms of service, we adopt a strict non-redistribution policy for all copyrighted material. 
The public release includes only derived, non-copyrighted metadata such as question–answer pairs, extracted factoids, aligned movie and recap-video time-stamps, YouTube identifiers, and links to publicly accessible resources (e.g., IMDb and OpenSubtitles). 
None of these elements contain audio, video frames, subtitle text, or other proprietary content; instead, they serve solely as pointers that allow researchers to reconstruct the data locally from legally obtained sources. 
This dataset is to be used for research purpose use only, and users remain responsible for complying with the copyright and platform terms associated with YouTube, IMDb, OpenSubtitles, and all other third-party sources.

Prior to the release of the dataset, we contacted all recap-video creators whose content is referenced by our metadata, informing them of the intended use and the nature of the release. 
None expressed objections, and we maintain an ongoing opt-out policy under which any creator may request removal of metadata referencing their video at any time. 
Through this design, \movierecapsqa supports open, reproducible scientific research while respecting copyright protections and the rights of all content creators involved.

%% file: supplementary/B-C.dataset-details.tex
\begin{table}
    \centering
    \small
    \resizebox{\columnwidth}{!}{  
    \begin{tabular}{p{\columnwidth}}
        \toprule
        \textbf{Instructions}: You are a helpful assistant who can extract atomic claims from a piece of text.\\
        \midrule
        You are trying to create a database of facts for the given text by extracting all atomic claims. To do so, you need to break down a sentence and extract as many fine-grained facts mentioned in the response.  Each fact should also be describing either one single event with necessary time and location information.  \\ \\
        You should focus on the named entities and numbers in the sentence and extract relevant information from the sentence. Recover pronouns, definite phrases (e.g., "the victims" or "the pope"), and so on. Each fact should be understandable on its own and require no additional context. I will provide you with the facts from the previous segment to use as context to do coreference resolution.  \\ \\
        All entities must be referred to by name but not pronoun. Use the name of entities rather than definite noun phrases (e.g., 'the teacher') whenever possible. If a definite noun phrase is used, be sure to add modifiers (e.g., a embedded clause, a prepositional phrase, etc.). Each fact must be situated within relevant temporal and location whenever needed. Keep each fact to one sentence with zero or at most one embedded clause. You do not need to justify what you extract. \\ \\
        
        Previous Segment's Facts: 
        \{ \textit{previous\_facts} \} \\
        Extract \textbf{atomic} fact from Text: \\
        \{\textit{text}\} \\
        \bottomrule
    \end{tabular}}
    \caption{\textbf{Prompt for Atomic Fact Extraction from Segments.}}
    \label{table:prompt_fact_extraction}
\end{table}

\section{Dataset Details -- Collection \& Generation}
\label{sec:appendix_dataset_creation_details}

In this section, we provide additional details on the construction of our dataset. As described in Section~\ref{sec:dataset}, we build \movierecapsqa around publicly available recap videos for movies that also have publicly accessible movie scripts. 
This design choice ensures that future extensions of the benchmark can seamlessly incorporate full movie scripts—which offer a richer and more accurate textual source than subtitles—without altering the core data pipeline. 

\paragraph{Movie Selection and Collection.}
We select movies using the open-source, fan-made scripts available on \href{https://imsdb.com}{IMSDb}. 
We crawl all available scripts and extract the associated movie titles, and then use the \href{https://developer.imdb.com}{official IMDb API} to obtain the corresponding movie metadata. 
We then obtain movie subtitles by searching and downloading the movie subtitles from the open source website, \href{https://www.opensubtitles.org/en/search/subs}{OpenSubtitles}.

\paragraph{Recap Video Collection.}
We search YouTube using the official API and retrieve the top five results for the query ``\textcolor{blue}{\bf recap video for \{movie\_name\}--\{movie\_release\_year\}}". 
We download the video, metadata, and audio for each YouTube video using their official API. 
We then extract the movie title by applying OCR to sampled 10 frame within the first ten seconds; for certain channels, we instead use a set of rule-based heuristics for title extraction.
This step is necessary because many recap creators intentionally avoid including the movie name in the title or description to encourage viewer engagement. 
After extracting candidate titles, we manually verify the alignment between each recap video and its corresponding movie to ensure that no mismatches occur.
Lastly, we obtain the recap video captions through a paid service, \href{https://www.rev.ai}{Rev.ai}.

\paragraph{Recap Video Scene Segmentation}
As described in Section~\ref{sec:dataset}, we segment each recap video using SceneDetect \citep{islam2023efficientmoviescenedetection}. This step is applied to the video component only. The scene segmenter operates by first detecting individual shots and then grouping consecutive shots whose visual distance is below a threshold, using the tuned SceneDetect embedding model. We use the publicly available implementation provided by authors of \citet{islam2023efficientmoviescenedetection}.

\paragraph{QA and Fact Generation}
The dataset is constructed through a three-stage, prompt-based generation pipeline. First, we extract atomic facts from each segmented portion of the recap video, prompt found in Table~\ref{table:prompt_fact_extraction}. Next, we generate question–answer pairs grounded in one or more of these facts, prompt found in Table~\ref{table:prompt_QA_generation}. Finally, we simplify the questions to increase their difficulty and reduce dependence on explicit character names, prompt found in Table~\ref{table:prompt_QA_simplification}. We provide more examples in Table~\ref{table:examples}. 

\begin{table}
    \centering
    \small
    \resizebox{\columnwidth}{!}{  
    \begin{tabular}{p{\columnwidth}}
        \toprule
        \textbf{Instructions}\\
        \midrule
        Generate question with their answers from the atomic facts generated. Make sure the answers are obtained from the atomic facts. Create complex questions that would require the use of one or multiple facts.
        
        Segment's Facts: 
        \{ \textit{segment\_facts} \} \\
        \bottomrule
    \end{tabular}
    }
    \caption{\textbf{Prompt for QA Generation.}}
    \label{table:prompt_QA_generation}
\end{table}

\begin{table}
    \centering
    \small
    \resizebox{\columnwidth}{!}{  
    \begin{tabular}{p{\columnwidth}}
        \toprule
        \textbf{Instructions}\\
        \midrule
Simplify and make a list of questions with their answers more ambiguous by removing unnecessary conditions or specifics. Your goal is to create a version of each question where key details are generalized or removed. If you remove a condition from the question, make sure that the answer reflects that change. Do not change ambiguous the answer. \\
\textbf{\#\# Steps:} \\
\hspace*{1em}1. \textbf{Identify Key Details}: For each question, pinpoint specific details or conditions that could be removed or generalized.\\
\hspace*{1em}2. \textbf{Generalize or Remove}: Simplify by either making the question more general or eliminating specific conditions, without losing the main intent.\\
\hspace*{1em}3. \textbf{Ensure Ambiguity}: Aim to increase the ambiguity in each question so that it can be interpreted in multiple ways.\\
\hspace*{1em}4. \textbf{Align Answers}: Adjust the answers as needed to reflect the removal or generalization of conditions. \\
\hspace*{1em}5. \textbf{Remove Names}: Adjust the questions by removing all names. \\
\textbf{\#\# Output Format:} \\
Provide simplified versions of the original questions, each on a new line, ensuring that answers are adjusted accordingly. Retain the order of the original list. It should be a list of json objects. \\
\textbf{\#\# Example:} \\
\hspace*{2em}Original Question:  What did Alex use as bait on the day his son was born? \\
\hspace*{2em}Original Answer: Alex used his gold wedding ring as bait. \\
\hspace*{2em}Simplified Questions: What did he use as bait on an important day? \\
\hspace*{2em}Simplified Answer: He used something precious as bait. \\ \\

        \{ \textit{question\_answer\_pair} \} \\
        \bottomrule
    \end{tabular}}
    \caption{\textbf{Prompt for QA Simplification.}}
    \label{table:prompt_QA_simplification}
\end{table}

\section{Dataset Details -- Alignment}
\label{sec:appendix_dataset_alignment}

\begin{figure*}[htb!]
    \centering
    \begin{tabular}{cc}
        \subfloat[Alignment between recap video  \textit{7zJ0nvgWpgk} and the movie \textit{Year One}]{%
            \includegraphics[width=0.45\linewidth]{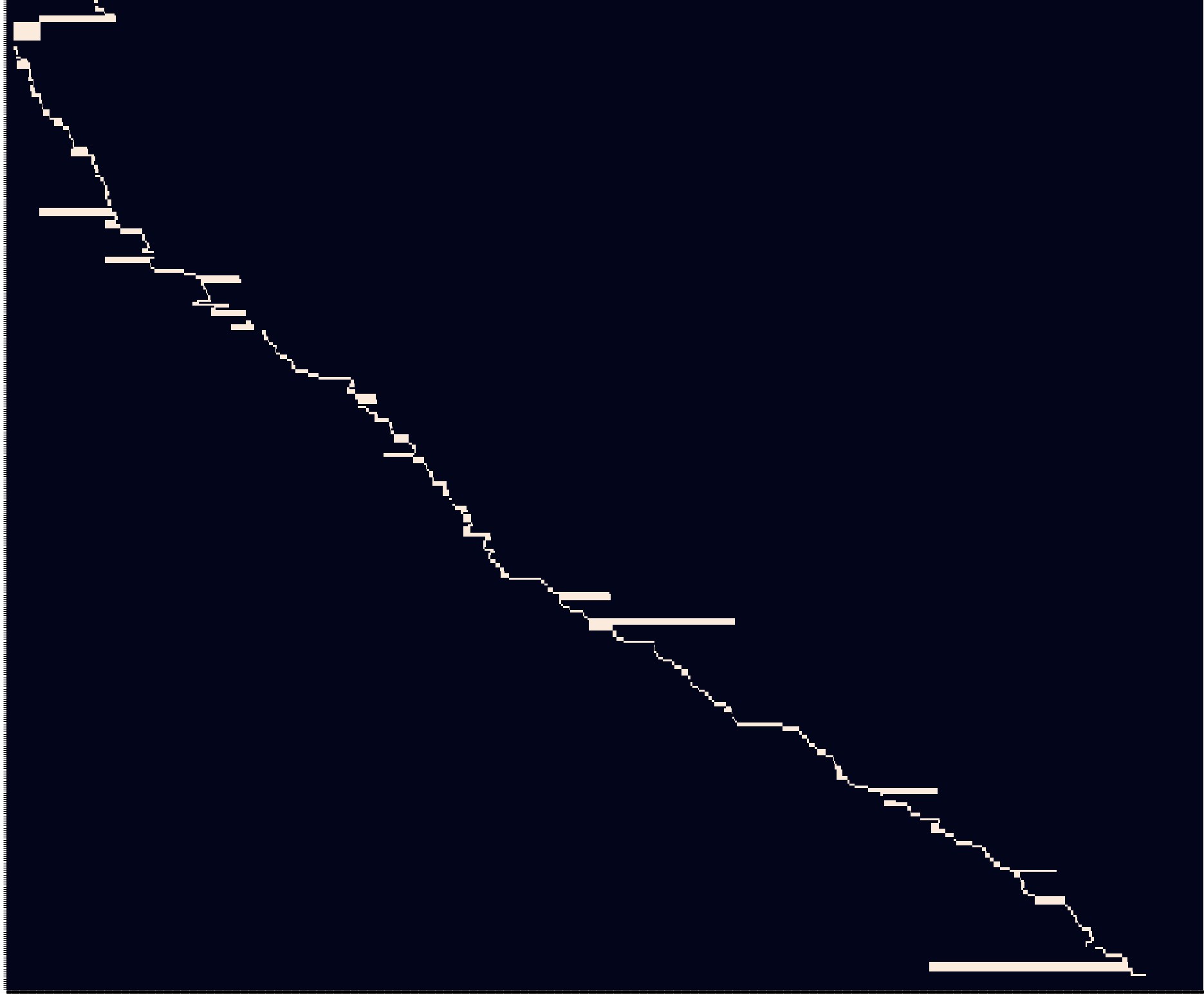}
            \label{subfigure:alignment.7zJ0nvgWpgk}
        } &
       \subfloat[Alignment between recap video \textit{pB6ULQIgmfg} and the movie \textit{Mission Impossible}]{%
            \includegraphics[width=0.45\linewidth]{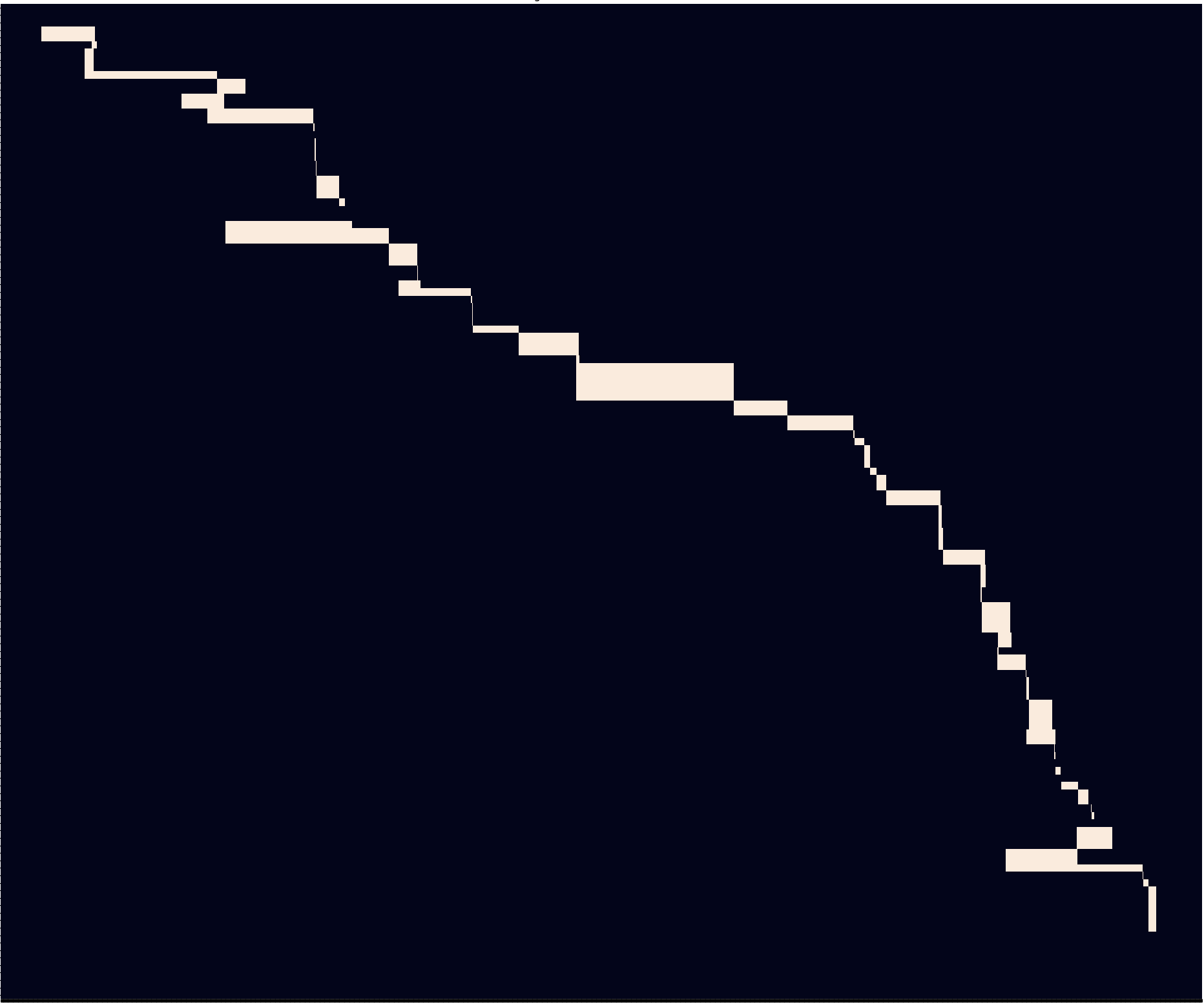}
            \label{subfigure:alignment.pB6ULQIgmfg}
        }
    \end{tabular}
    \caption{\textbf{Alignment Between Recap Videos and Full Movies.} We show alignment examples for selected videos in the dataset. Each plot maps recap-video time-stamps (y-axis) to movie time-stamps (x-axis) using our segment-shot similarity procedure. While most alignments follow a near-diagonal structure, indicating chronological correspondence, recap videos occasionally reorder scenes for narrative flow (e.g., character introductions), resulting in local misalignments.}
    \label{figure:alignment}
\end{figure*}

A distinctive property of recap videos is that their visual content alone functions as a condensed summary of the movie. 
Recap creators typically stitch together the key shots that convey the major plot points and narrative transitions. 
As a result, it becomes possible to establish a meaningful alignment between the time-frames in the recap video and those in the full movie. 
This alignment, in turn, enables efficient matching between recap time-frames and movie subtitles, since subtitles are inherently time-stamped. \footnote{We obtain the movie videos using Amazon Prime subscriptions.}

For optimal alignment, we first detect all shots in both the full movie using a standard shot-detection pipeline. We encode each shot and segment by extracting SlowFast-50 embeddings from the first and last three seconds. 
To establish correspondences, we compute the maximum softmax similarity between the start–end embeddings of each recap shot and those of the movie, selecting the highest-confidence matches as candidate alignments while enforcing continuity and chronological consistency. Figure~\ref{figure:alignment} shows example alignments from the dataset, where the x-axis denotes movie time-stamps and the y-axis denotes recap-video time-stamps.

However, because recap videos often reorder shots to maintain narrative flow, the resulting alignment is not always strictly chronological and can occasionally drift. This effect is visible in Figure~\ref{subfigure:alignment.7zJ0nvgWpgk}, where the alignment briefly moves out of order near the top-left region, reflecting cases in which recap creators introduce characters or events earlier than they appear in the movie. To address such issues, we apply a series of heuristic filtering and smoothing techniques to enforce temporal consistency and remove implausible matches.\footnote{All alignment code and heuristics used in this process will be released with the dataset.}

An additional observation about recap videos is that some movie scenes require disproportionately longer summaries. As shown in Figure~\ref{subfigure:alignment.pB6ULQIgmfg}, a long stretch along the y-axis (recap time) aligns to a much shorter segment on the x-axis (movie time), indicating that a small cluster of shots in the movie contains dense, plot-critical information that warrants a more detailed recap. This behavior is especially common in action or spy movies (e.g., \textit{Mission: Impossible}), where pivotal sequences—such as heists or reveal moments—contain numerous important events compressed into a short timeframe.

%% file: supplementary/D.problem-setting.tex
\section{Alternative Problem Settings}
\label{sec:appendix_alternate}

Primary in this study, we evaluated the performance of MLLMs on \movierecapsqa benchmark using the recap segment, $s_r \subset v_r$ (typically $\sim 73$ seconds) as the video context to the questions. 
However, the benchmark provides deeper video--text alignments that enable several alternative and increasingly challenging evaluation modes.

\paragraph{Alignment Units.}
For each QA pair, the benchmark supplies: (1) the full recap video $v_r$ (8--15 minutes), (2) the specific recap segment $s_r$ from which the question is constructed, (3) the aligned movie segment $s_m$ from the original film, and (4) the complete movie $v_m$. 
Textual components are also aligned across levels: the movie subtitles $u_m$ align with $s_m$. 
The pair is also combined with the fact set $F = \{f_1, \ldots, f_k\}$ obtained from $s_r$. 
Each datapoint in \movierecapsqa provides a structured multimodal tuple
\[
\{\text{Question}, \text{Answer}, v_r,\, s_r,\, v_m,\, s_m,\, u_m,\, F\},
\]
which supports multiple interchangeable input configurations.

\paragraph{Default Recap-Segment QA.}
The main benchmark setting evaluates models using only the recap segment:
\[
\langle s_r,\, u_m,\, \text{Question} \rangle \rightarrow \text{Answer}.
\]
This is the current setting designed in this study as it is the most cost effictive.

\paragraph{Movie-Segment QA.}
The benchmark also includes the movie segment $s_m$ temporally aligned to the same narrative moment as $s_r$. This allows a formulation where models answer from original film footage rather than recap edits:
\[
\langle s_m,\, u_m,\, \text{Question} \rangle \rightarrow \text{Answer}.
\]
This setting isolates how models behave when provided full video context which differs in from the compressed video recap summaries. 

\paragraph{Full-Recap QA.}
Instead of using a short clip, models may be evaluated on the entire recap video:
\[
\langle v_r,\, u_m,\,  \text{Question} \rangle \rightarrow \text{Answer}.
\]
This introduces long-range multimodal reasoning and requires models to track character arcs, causality, and scene transitions across a 10–15 minute video summary.

\paragraph{Full-Movie QA.}
The benchmark further supports a long-video setting in which the full movie $v_m$ and its subtitles $u_m$ are used as input:
\[
\langle v_m,\, u_m,\,  \text{Question} \rangle \rightarrow \text{Answer}.
\]
This is the most challenging configuration, demanding temporal grounding and narrative understanding across an entire 1.5–2.5 hour film.

\paragraph{Summary.}
These four configurations---recap segment, movie segment, full recap, and full movie---provide a continuum of evaluation difficulty, all derived from the same aligned QA pairs. 
While the benchmark’s default setting emphasizes short, efficient VideoQA, its multi-granular alignment structure enables a broad family of alternative tasks including long-video comprehension, cross-modal alignment (recap $\leftrightarrow$ movie), scene retrieval, and fact verification over extended multimodal context. This flexibility allows \movierecapsqa to serve not only as a short-video QA benchmark but also as a testbed for long video context.

%% file: supplementary/E.question-categorization-and-modality.tex
\begin{table}
    \centering
    \small
    \resizebox{\columnwidth}{!}{
    \begin{tabular}{p{\columnwidth}}
        \toprule
        \textbf{Instructions}: You are a helpful assistant who classifies movie questions into semantic categories. \\
        \midrule

        Your task is to analyze each question from the movie ``\{MOVIE\_NAME\}'' and assign it to \textbf{exactly one} semantic category. The categories capture different dimensions of reasoning required to answer the question. \\ \\
        \textbf{TEMP (Temporal)}: questions involving time, sequence of events, chronology, duration, or ordering of actions. \\
        \textbf{CRD (Character and Relationship Dynamics)}: questions about character motivations, emotions, intentions, interactions, or interpersonal relationships. \\
        \textbf{NPA (Narrative and Plot Analysis)}: questions about story structure, plot developments, causal reasoning, or narrative elements. \\
        \textbf{STA (Setting and Technical Analysis)}: questions focusing on location, environment, staging, cinematography, or production-related elements. \\
        \textbf{TH (Thematic Exploration)}: questions addressing themes, symbolism, messages, or deeper conceptual meanings. \\ \\
        For each provided question, output only its corresponding category label. \\ \\
        \textbf{Questions}: \{questions\} \\ \\
        \textbf{Categories}: \\
        \bottomrule
    \end{tabular}
    }
    \caption{\textbf{Prompt for Movie Question Categorization.}}
    \label{table:prompt_categorization}
\end{table}

\section{Question Categorization and Modality Prompt Templates}
\label{sec:appendix_question_cats_modality}

We categorize each question into one of five semantic reasoning categories, using the prompt shown in Table~\ref{table:prompt_categorization}. The categories are:

\begin{itemize}
    \item \textbf{NPA} (Narrative and Plot Analysis): questions concerning events, causality, and overall story progression.
    \item \textbf{CRD} (Character and Relationship Dynamics): questions involving character traits, emotions, intentions, or interpersonal relationships.
    \item \textbf{TH} (Thematic Exploration): questions about themes, moral lessons, symbolism, or overarching narrative messages.
    \item \textbf{STA} (Setting and Technical Attributes): questions focused on locations, visual style, cinematography, or production-related elements.
    \item \textbf{TEMP} (Temporal Reasoning): questions that involve ordering, duration, or timing of events within the segment.
\end{itemize}

In addition, we annotate each question with a modality type. To determine this, we provide the model with the question, the associated dialogue, and a visual scene description, stored collectively as \verb|context_pairs|. The prompt for this task appears in Table~\ref{table:prompt_modality}, and the modality types are:

\begin{itemize}
    \item \textbf{Dialogue-based}: answerable solely from dialogue (spoken lines or subtitles), without requiring visual cues.
    \item \textbf{Scene-based}: requiring information from the visual scene (characters, actions, objects, locations) that cannot be inferred from dialogue alone.
    \item \textbf{Multimodal}: requiring both dialogue and visual information, where neither modality alone is sufficient.
\end{itemize}

\begin{table}
    \centering
    \small
    \resizebox{\columnwidth}{!}{
    \begin{tabular}{p{\columnwidth}}
        \toprule
        \textbf{Instructions}:  You are a helpful assistant who classifies movie questions into semantic categories. \\
        \midrule
        Analyze the following questions from the movie ``\{MOVIE\_NAME\}'' and classify each one into exactly one modality type. \\ \\
        \textbf{Dialogue-based}: can be answered solely from dialogue (subtitles or spoken lines), without requiring any visual information. \\
        \textbf{Scene-based}: requires visual scene information (characters, actions, objects, locations) and cannot be answered from dialogue alone. \\
        \textbf{Multimodal}: requires both dialogue and visual information; neither modality alone is sufficient. \\ \\
        Here are the questions to classify:\\
        \bottomrule
    \end{tabular}
    }
    \caption{\textbf{Prompt for Question Modality Classification (Dialogue, Scene, Multimodal).}}
    \label{table:prompt_modality}
\end{table}

%% file: supplementary/F.evaluation-prompts.tex
\section{Evaluation Metric Prompts}
\label{sec:appendix_evalaution_prompts}

In this section, we provide the full prompts used to evaluate model responses in our benchmark. 
All evaluations are executed using the OpenAI Batch API, which allows us to scale the assessment of thousands of model outputs in a cost-efficient and environmentally responsible manner.

\paragraph{Atomic Claim Extraction.}  
Given a model-generated answer, we first extract its underlying atomic claims using the prompt shown in Table~\ref{table:prompt_atomic_extraction_model}.  
This step decomposes the model’s answer into fine-grained, verifiable units of meaning, enabling consistent downstream factuality and relevance evaluation. This is inspired by work on factuality evaluation in text question-answering.

\paragraph{Factuality Evaluation.}  
To evaluate factual correctness, we supply the evaluator with: (i) all atomic facts extracted from the corresponding video segment, (ii) the model’s answer claims, (iii) the question, and (iv) the aligned SRT dialogue.  
This design reflects the core principle of factuality: a model’s answer must not introduce information that contradicts, hallucinates beyond, or misrepresents the input evidence.  
Because the segment-level facts serve as a textual representation of the visual content, and the SRT dialogue captures additional narrative cues, these two sources together provide comprehensive grounding for judging factual accuracy.  
The exact factuality evaluation prompt is provided in Table~\ref{table:prompt_factuality_evaluation}.

\paragraph{Relevance Evaluation.}  
For relevance, we evaluate whether each extracted claim meaningfully contributes to answering the user’s question.  
We provide the evaluator with: (i) the question, (ii) the model’s answer claims, (iii) the SRT dialogue, and (iv) the \emph{aligned facts used to generate that question}.  
These aligned facts encode the semantic intent of the question, allowing the evaluator to determine whether a claim ``belongs" to the same underlying evidence that motivated the question.  
Unlike factuality, correctness plays no role here—a claim can be factually wrong yet still relevant if it attempts to answer the question.  
The full relevance evaluation prompt is shown in Table~\ref{table:prompt_relevance_evaluation}.

\begin{table}
    \centering
    \small
    \resizebox{\columnwidth}{!}{
    \begin{tabular}{p{\columnwidth}}
        \toprule
        \textbf{Instructions}: You are a helpful assistant who can extract atomic claims from a piece of text. \\
        \midrule

        You are trying to verify how factual a response to a question or request is. To do so, you must break down the model's answer into as many fine-grained, atomic facts as possible. Each fact must describe a single event, state, or relation, including necessary temporal or location information when relevant. \\ \\

        Focus on named entities and numbers, and extract all relevant information expressed in the sentence. Do \textbf{not} extract claims from the question itself; the question serves only as context to resolve pronouns, definite noun phrases (e.g., “the victims”, “the pope”), and other referring expressions. Each fact must be understandable on its own, without requiring additional context. \\ \\

        All entities should be referred to by explicit name rather than pronoun. When using definite noun phrases, include modifiers (e.g., embedded clauses, prepositional phrases) to ensure specificity. Each fact should be one sentence long, with zero or at most one embedded clause. You do not need to justify the extracted facts. \\ \\

        Extract \textbf{atomic} facts. \\ \\

        \textbf{Question}: \{question\} \\
        \textbf{Model Answer}: \{answer\} \\ \\
        \textbf{Facts}: \\

        \bottomrule
    \end{tabular}
    }
    \caption{\textbf{Prompt for Atomic Fact Extraction from Model Answers.}}
    \label{table:prompt_atomic_extraction_model}
\end{table}

%% file: supplementary/G.coherence-study.tex
\section{Evaluation Metric -- Coherence}
\label{sec:appendix_coherence}
Beyond factuality and relevance, coherence is frequently used as an auxiliary evaluation dimension in text-based question answering, as seen in metrics such as G-Eval Fluency and HELMET Fluency. Motivated by this, we also incorporate a coherence assessment in our benchmark.

To evaluate coherence, we provide the evaluator with: (i) the model’s extracted answer claims, and (ii) the question. 
Coherence here measures the internal logical consistency of the response: a coherent answer should not contain claims that contradict one another or repeat the same information unnecessarily. 
The full coherence evaluation prompt is provided in Table~\ref{table:prompt_coherence_evaluation}.

However, unlike long-form text generation tasks, coherence is significantly less informative in open-ended VideoQA. 
Answers in VideoQA are typically short and contain very few distinct claims, which greatly limits the possibility of internal contradictions. 
Consequently, coherence scores exhibit extremely low variance across models. 
We report these results on the 118 manually evaluated questions in Table~\ref{table:coherence_scores}, where the near-uniform scores confirm that coherence is not a meaningful discriminative dimension for short-form VideoQA responses. 
For this reason, while we compute coherence for completeness, it does not play a substantial role in evaluating model performance in our setting.

\begin{table}[thb!]
\centering
\small
\begin{tabular}{lc}
\toprule
\textbf{Model}  & \textbf{Coherence Evaluation} \\
\midrule

LLaVA-NeXT-Video      
& $4.79_{\pm 0.41}$ \\

MiniCPM-o             
& $4.75_{\pm 0.28}$ \\

Qwen                  
& $4.82_{\pm 0.28}$ \\

Amazon Nova           
& $4.71_{\pm 0.40}$ \\

Claude                
& $4.65_{\pm 0.40}$ \\

Gemini 2.5 Flash      
& $4.81_{\pm 0.37}$ \\

GPT-4o                
& $4.77_{\pm 0.36}$ \\

\midrule
\rowcolor{blue!10}
Avg. Human*  
& $4.86_{\pm 0.31}$ \\

\rowcolor{blue!10}
Best Human* 
& $5.00_{\pm 0.00}$ \\
\bottomrule
\end{tabular}
\caption{
\textbf{Coherence Scores (1–5).} 
Mean $\pm$ variance coherence scores for all models and human annotators, rounded to two decimal places. 
The consistently low variance confirms that coherence is not a discriminative metric for short-form VideoQA.
}
\label{table:coherence_scores}
\end{table}


%% file: supplementary/H.detailed-ablation-experiments.tex
\section{Detailed Ablation Tables}
\label{sec:appendix_detailed_ablation_results}
In this section, we provide the complete ablation results for each model under all evaluation settings. 
Table~\ref{tab:model_relevance_types_categories} reports the relevance scores across question types and categories, while Table~\ref{tab:model_factuality_types_categories} presents the corresponding factuality scores. 
These tables complement the main results by showing how each model behaves when provided with different input modalities and ablated forms of the video–text context.

\paragraph{Summary of Model Behaviors.}
Tables~\ref{tab:model_relevance_types_categories}--\ref{tab:model_factuality_types_categories} reveal consistent trends across both relevance and factuality. Proprietary models (GPT-4o, Claude~3.5~Sonnet, Amazon Nova Lite) outperform open-source systems across nearly all question types and categories when given full context. Dialogue-based questions are the easiest for all models, while scene-based and multimodal questions expose clear gaps in visual grounding, especially for open-source models.

\paragraph{Effect of Input Ablations.}
Dialogue-only inputs yield the strongest gains for most models—often surpassing the full-context baseline by large margins. GPT-4o and Claude~3.5 improve by up to \(+0.4\)–\(0.7\) in both relevance and factuality, indicating that subtitles carry the dominant grounding signal. In contrast, frames-only inputs provide limited benefit: Qwen2.5VL is the only model that consistently improves under visual-only conditions, particularly for temporal and spatial reasoning. For other models, removing subtitles significantly degrades performance.

\paragraph{Category-Level Observations.}
CRD, NPA, and TH categories benefit most from dialogue, reflecting their reliance on explicit narrative cues. STA and TEMP remain the most challenging: even strong models show reduced factual grounding, and multimodal fusion rarely helps. Notably, Gemini-2.5-Flash exhibits the weakest temporal grounding (e.g., TEMP factuality 2.53), while Qwen2.5VL achieves the strongest temporal improvements under frames-only input.

\paragraph{Overall.}
The evaluation highlights three core findings: (i) subtitles dominate model grounding on recap-derived QA; (ii) visual understanding remains model-dependent and uneven; and (iii) multimodal fusion remains an open challenge, with models frequently performing better when one modality is suppressed.

\input{tables/ablation_results}

%% file: tables/ablation_results.tex
\begin{table*}[t]
\centering
\small
\setlength{\tabcolsep}{4pt}
\begin{tabular}{lccccccccc}
\toprule
 & \multicolumn{3}{c}{\textbf{Question Types}} & \multicolumn{5}{c}{\textbf{Question Categories}} \\
\cmidrule(lr){2-4} \cmidrule(lr){5-9}
\textbf{Model} & \textbf{Dialogue} & \textbf{Scene} & \textbf{Multimodal} & \textbf{CRD} & \textbf{NPA} & \textbf{STA} & \textbf{TEMP} & \textbf{TH} \\
\midrule
\rowcolor{blue!5}
\textbf{LLaVA-NeXT-Video} &3.36 & 3.35 & 3.33 & 3.30 & 3.31 & 3.37 & 3.54 & 3.52 \\

(only frames) 
& 3.37{\color{BetterGreen}↑} & 3.41{\color{BetterGreen}↑} & 3.43{\color{BetterGreen}↑} 
& 3.45{\color{BetterGreen}↑} & 3.32{\color{BetterGreen}↑} & 3.41{\color{BetterGreen}↑} 
& 3.43{\color{BetterRed}↓} & 3.53{\color{BetterGreen}↑} \\

(only dialogue) 
& 3.59{\color{BetterGreen}↑} & 3.40{\color{BetterGreen}↑} & 3.48{\color{BetterGreen}↑}
& 3.56{\color{BetterGreen}↑} & 3.45{\color{BetterGreen}↑} & 3.42{\color{BetterGreen}↑}
& 3.43{\color{BetterRed}↓} & 3.54{\color{BetterGreen}↑} \\

\rowcolor{blue!5}
\textbf{Mini-CPM-o} &3.54 & 3.55 & 3.52 & 3.52 & 3.50 & 3.56 & 3.66 & 3.74 \\

(only frames)
& 3.70{\color{BetterGreen}↑} & 3.72{\color{BetterGreen}↑} & 3.71{\color{BetterGreen}↑}
& 3.71{\color{BetterGreen}↑} & 3.65{\color{BetterGreen}↑} & 3.68{\color{BetterGreen}↑}
& 3.72{\color{BetterGreen}↑} & 3.84{\color{BetterGreen}↑} \\

(only dialogue)
& 3.86{\color{BetterGreen}↑} & 3.65{\color{BetterGreen}↑} & 3.82{\color{BetterGreen}↑}
& 3.81{\color{BetterGreen}↑} & 3.73{\color{BetterGreen}↑} & 3.72{\color{BetterGreen}↑}
& 3.74{\color{BetterGreen}↑} & 4.00{\color{BetterGreen}↑} \\

\rowcolor{blue!5}
\textbf{Qwen2.5VL} &3.93 & 3.69 & 3.72 & 3.78 & 3.75 & 3.80 & 3.90 & 3.91 \\

(only frames)
& 3.92{\color{BetterRed}↓} & 3.88{\color{BetterGreen}↑} & 3.87{\color{BetterGreen}↑}
& 3.88{\color{BetterGreen}↑} & 3.80{\color{BetterGreen}↑} & 3.88{\color{BetterGreen}↑}
& 3.95{\color{BetterGreen}↑} & 4.15{\color{BetterGreen}↑} \\

(only dialogue)
& 3.52{\color{BetterRed}↓} & 3.27{\color{BetterRed}↓} & 3.51{\color{BetterRed}↓}
& 3.49{\color{BetterRed}↓} & 3.45{\color{BetterRed}↓} & 3.17{\color{BetterRed}↓}
& 3.42{\color{BetterRed}↓} & 3.65{\color{BetterRed}↓} \\

\rowcolor{blue!5}
\textbf{Amazon Nova Lite} &4.12 & 3.82 & 3.99 & 3.97 & 3.95 & 3.81 & 3.94 & 4.23 \\

(only frames)
& 3.81{\color{BetterRed}↓} & 3.53{\color{BetterRed}↓} & 3.77{\color{BetterRed}↓}
& 3.78{\color{BetterRed}↓} & 3.66{\color{BetterRed}↓} & 3.74{\color{BetterRed}↓}
& 3.39{\color{BetterRed}↓} & 3.73{\color{BetterRed}↓} \\

(only dialogue) 
& -- & -- & -- & -- & -- & -- & -- & -- \\

\rowcolor{blue!5}
\textbf{Claude 3.5 Sonnet} &3.88 & 3.71 & 3.83 & 3.86 & 3.72 & 3.61 & 3.99 & 3.82 \\

(only frames)
& 3.48{\color{BetterRed}↓} & 3.68{\color{BetterRed}↓} & 3.68{\color{BetterRed}↓}
& 3.58{\color{BetterRed}↓} & 3.60{\color{BetterRed}↓} & 3.62{\color{BetterGreen}↑}
& 3.77{\color{BetterRed}↓} & 3.54{\color{BetterRed}↓} \\

(only dialogue)
& 4.09{\color{BetterGreen}↑} & 3.84{\color{BetterGreen}↑} & 4.07{\color{BetterGreen}↑}
& 4.08{\color{BetterGreen}↑} & 3.99{\color{BetterGreen}↑} & 3.77{\color{BetterGreen}↑}
& 3.87{\color{BetterRed}↓} & 4.10{\color{BetterGreen}↑} \\

\rowcolor{blue!5}
\textbf{Gemini-2.5-Flash} &3.66 & 3.45 & 3.67 & 3.67 & 3.58 & 3.38 & 3.41 & 3.62 \\

(only frames)
& 3.55{\color{BetterRed}↓} & 3.38{\color{BetterRed}↓} & 3.51{\color{BetterRed}↓}
& 3.49{\color{BetterRed}↓} & 3.47{\color{BetterRed}↓} & 3.38{\color{BetterGreen}↑}
& 3.44{\color{BetterGreen}↑} & 3.58{\color{BetterRed}↓} \\

(only dialogue)
& 4.07{\color{BetterGreen}↑} & 3.77{\color{BetterGreen}↑} & 4.03{\color{BetterGreen}↑}
& 4.05{\color{BetterGreen}↑} & 3.95{\color{BetterGreen}↑} & 3.67{\color{BetterGreen}↑}
& 3.73{\color{BetterGreen}↑} & 4.02{\color{BetterGreen}↑} \\

\rowcolor{blue!5}
\textbf{GPT-4o} &3.71 & 3.55 & 3.84 & 3.78 & 3.73 & 3.32 & 3.59 & 3.76 \\

(only frames)
& 3.83{\color{BetterGreen}↑} & 3.99{\color{BetterGreen}↑} & 3.93{\color{BetterGreen}↑}
& 3.87{\color{BetterGreen}↑} & 3.91{\color{BetterGreen}↑} & 3.82{\color{BetterGreen}↑}
& 3.89{\color{BetterGreen}↑} & 3.91{\color{BetterGreen}↑} \\

(only dialogue)
& 4.20{\color{BetterGreen}↑} & 3.82{\color{BetterGreen}↑} & 4.13{\color{BetterGreen}↑}
& 4.16{\color{BetterGreen}↑} & 4.05{\color{BetterGreen}↑} & 3.80{\color{BetterGreen}↑}
& 3.85{\color{BetterGreen}↑} & 4.12{\color{BetterGreen}↑} \\

\bottomrule
\end{tabular}
\caption{\textbf{Model performance (mean relevance) across question types and categories.} Arrows indicate increase/decrease relative to the full model baseline.}
\label{tab:model_relevance_types_categories}
\end{table*}

\begin{table*}[t]
\centering
\small
\setlength{\tabcolsep}{4pt}

\definecolor{BetterGreen}{RGB}{34,139,34} 
\definecolor{BetterRed}{RGB}{178,34,34}   

\begin{tabular}{lccccccccc}
\toprule
 & \multicolumn{3}{c}{\textbf{Question Types}} & \multicolumn{5}{c}{\textbf{Question Categories}} \\
\cmidrule(lr){2-4} \cmidrule(lr){5-9}
\textbf{Model} & \textbf{Dialogue} & \textbf{Scene} & \textbf{Multimodal} & \textbf{CRD} & \textbf{NPA} & \textbf{STA} & \textbf{TEMP} & \textbf{TH} \\
\midrule

\rowcolor{blue!5}
\textbf{LLaVA-NeXT-Video} 
&2.99 \ \  & 2.88 & 2.88 & 2.99 & 2.90 & 2.65 & 3.04 & 2.78 \\

(only frames)
& 2.69{\color{BetterRed}↓} 
& 2.82{\color{BetterRed}↓} 
& 2.78{\color{BetterRed}↓}
& 2.79{\color{BetterRed}↓}
& 2.77{\color{BetterRed}↓}
& 2.64{\color{BetterRed}↓}
& 2.73{\color{BetterRed}↓}
& 2.64{\color{BetterRed}↓} \\

(only dialogue)
& 3.14{\color{BetterGreen}↑}
& 2.93{\color{BetterGreen}↑}
& 3.02{\color{BetterGreen}↑}
& 3.14{\color{BetterGreen}↑}
& 3.02{\color{BetterGreen}↑}
& 2.81{\color{BetterGreen}↑}
& 2.92{\color{BetterRed}↓}
& 2.91{\color{BetterGreen}↑} \\

\rowcolor{blue!5}
\textbf{Mini-CPM-o} 
&3.15 & 3.00 & 3.09 & 3.14 & 3.10 & 2.76 & 3.02 & 3.02 \\

(only frames)
& 3.11{\color{BetterRed}↓}
& 3.10{\color{BetterGreen}↑}
& 3.07{\color{BetterRed}↓}
& 3.13{\color{BetterRed}↓}
& 3.09{\color{BetterRed}↓}
& 2.84{\color{BetterGreen}↑}
& 3.10{\color{BetterGreen}↑}
& 2.87{\color{BetterRed}↓} \\

(only dialogue)
& 3.37{\color{BetterGreen}↑}
& 3.17{\color{BetterGreen}↑}
& 3.35{\color{BetterGreen}↑}
& 3.41{\color{BetterGreen}↑}
& 3.37{\color{BetterGreen}↑}
& 2.94{\color{BetterGreen}↑}
& 3.13{\color{BetterGreen}↑}
& 3.07{\color{BetterGreen}↑} \\

\rowcolor{blue!5}
\textbf{Qwen2.5VL} 
&3.50 & 3.28 & 3.35 & 3.42 & 3.40 & 3.07 & 3.39 & 3.27 \\

(only frames)
& 3.42{\color{BetterRed}↓}
& 3.49{\color{BetterGreen}↑}
& 3.41{\color{BetterGreen}↑}
& 3.43{\color{BetterGreen}↑}
& 3.43{\color{BetterGreen}↑}
& 3.36{\color{BetterGreen}↑}
& 3.45{\color{BetterGreen}↑}
& 3.29{\color{BetterGreen}↑} \\

(only dialogue)
& 3.10{\color{BetterRed}↓}
& 3.10{\color{BetterRed}↓}
& 3.24{\color{BetterRed}↓}
& 3.21{\color{BetterRed}↓}
& 3.25{\color{BetterRed}↓}
& 2.58{\color{BetterRed}↓}
& 2.92{\color{BetterRed}↓}
& 3.16{\color{BetterRed}↓} \\

\rowcolor{blue!5}
\textbf{Amazon Nova Lite} 
&3.73 & 3.35 & 3.58 & 3.59 & 3.60 & 3.15 & 3.51 & 3.37 \\

(only frames)
& 2.98{\color{BetterRed}↓}
& 2.78{\color{BetterRed}↓}
& 3.17{\color{BetterRed}↓}
& 3.22{\color{BetterRed}↓}
& 3.02{\color{BetterRed}↓}
& 2.62{\color{BetterRed}↓}
& 2.69{\color{BetterRed}↓}
& 2.87{\color{BetterRed}↓} \\

(only dialogue)
& -- & -- & -- & -- & -- & -- & -- & -- \\

\rowcolor{blue!5}
\textbf{Claude 3.5 Sonnet} 
&3.69 & 3.17 & 3.58 & 3.65 & 3.42 & 3.12 & 3.30 & 3.44 \\

(only frames)
& 3.19{\color{BetterRed}↓}
& 3.15{\color{BetterRed}↓}
& 3.24{\color{BetterRed}↓}
& 3.23{\color{BetterRed}↓}
& 3.25{\color{BetterRed}↓}
& 2.81{\color{BetterRed}↓}
& 3.19{\color{BetterRed}↓}
& 3.05{\color{BetterRed}↓} \\

(only dialogue)
& 4.17{\color{BetterGreen}↑}
& 3.93{\color{BetterGreen}↑}
& 4.21{\color{BetterGreen}↑}
& 4.17{\color{BetterGreen}↑}
& 4.13{\color{BetterGreen}↑}
& 3.78{\color{BetterGreen}↑}
& 3.95{\color{BetterGreen}↑}
& 4.18{\color{BetterGreen}↑} \\

\rowcolor{blue!5}
\textbf{Gemini-2.5-Flash} 
&3.34 & 2.65 & 3.03 & 3.15 & 3.00 & 2.57 & 2.53 & 3.16 \\

(only frames)
& 2.99{\color{BetterRed}↓}
& 2.78{\color{BetterGreen}↑}
& 2.87{\color{BetterRed}↓}
& 2.94{\color{BetterRed}↓}
& 2.92{\color{BetterRed}↓}
& 2.50{\color{BetterRed}↓}
& 2.68{\color{BetterGreen}↑}
& 2.82{\color{BetterRed}↓} \\

(only dialogue)
& 4.03{\color{BetterGreen}↑}
& 3.77{\color{BetterGreen}↑}
& 4.11{\color{BetterGreen}↑}
& 4.13{\color{BetterGreen}↑}
& 4.00{\color{BetterGreen}↑}
& 3.48{\color{BetterGreen}↑}
& 3.66{\color{BetterGreen}↑}
& 3.87{\color{BetterGreen}↑} \\

\rowcolor{blue!5}
\textbf{GPT-4o} 
&3.76 & 3.43 & 3.66 & 3.73 & 3.64 & 3.10 & 3.58 & 3.55 \\

(only frames)
& 3.47{\color{BetterRed}↓}
& 3.54{\color{BetterGreen}↑}
& 3.58{\color{BetterRed}↓}
& 3.59{\color{BetterRed}↓}
& 3.57{\color{BetterRed}↓}
& 3.26{\color{BetterGreen}↑}
& 3.32{\color{BetterRed}↓}
& 3.26{\color{BetterRed}↓} \\

(only dialogue)
& 4.07{\color{BetterGreen}↑}
& 3.72{\color{BetterGreen}↑}
& 4.06{\color{BetterGreen}↑}
& 4.14{\color{BetterGreen}↑}
& 4.00{\color{BetterGreen}↑}
& 3.41{\color{BetterGreen}↑}
& 3.63{\color{BetterGreen}↑}
& 3.86{\color{BetterGreen}↑} \\

\bottomrule
\end{tabular}
\caption{\textbf{Model performance (mean factuality) across question types and categories.} Arrows indicate increase/decrease relative to the full model baseline.
}
\label{tab:model_factuality_types_categories}
\end{table*}

%% file: supplementary/Z.example-questions.tex
\begin{table*}
    \centering
    \small
    \resizebox{.9\textwidth}{!}{  
    \begin{tabular}{l p{.2\textwidth} p{.2\textwidth} p{.2\textwidth} l l }
        \toprule
         & \bf Verbose Question & \bf Question & \bf Answer & Category & Type \\
         \midrule

\emph{1} &
What indication is there that Jim is unaware of the circumstances in the city? &
What shows that Jim is unaware of the situation? &
Jim has no idea what happened, indicating his lack of knowledge about the situation in the city. &
NPA &
multimodal \\

\rowcolor{yellow!25}
\emph{2} &
What activities does he engage in as part of recalling his earlier years? &
What activities are linked to his fond memories? &
He reminisces about his past glory days as part of his daily routine. &
CRD &
dialogue-based \\

\emph{3} &
Where is he hoping to find other people? &
Where is he hoping to find others? &
He is hoping to find other people in the city. &
CRD &
scene-based \\

\rowcolor{yellow!25}
\emph{4} &
In what circumstances does Neo find himself after swallowing the red pill, and what does this reveal about his previous existence? &
What situation does someone find themselves in after making a choice, and what does this suggest about their past? &
After taking the red pill, Neo awakens in a pod among countless humans, revealing his previous life was a simulation. &
TH &
multimodal \\

\emph{5} &
Describe the sequence of events that occurs during the encounter between Deckard and Chris. &
What happens during the encounter between two characters? &
Deckard is ambushed by Chris, but he manages to kill her just as Roy returns. &
NPA &
multimodal \\

\rowcolor{yellow!25}
\emph{6} &
How does Dr. Chilton's behavior towards Clarice reflect the dynamics of their interaction? &
How does one person's behavior towards another reflect their interaction dynamics? &
Dr. Chilton's suggestive remarks show personal interest, contrasting with the professional setting and hinting at manipulation. &
CRD &
dialogue-based \\

\emph{7} &
What are the final moments witnessed by Deckard after Roy delivers his monologue? &
What does one character witness after a speech from another? &
Roy dies in front of Deckard, who watches silently as Gaff arrives and calls to him. &
CRD &
scene-based \\

\rowcolor{yellow!25}
\emph{8} &
What does Valentina believe is the only choice in the situation? &
What does Valentina think is the choice to make? &
Valentina states that the only choice is to help. &
CRD &
dialogue-based \\

\emph{9} &
What emotions does Andrew experience after stepping out of the party, and what does he do in the parking lot? &
What feelings does Andrew go through after leaving an event, and what does he do outside? &
Andrew cries as he cleans dirt off his camera after leaving the party. &
CRD &
multimodal \\

         \bottomrule
    \end{tabular}
    }
    \caption{Representative examples from the \movierecapsqa dataset. Each entry shows the question, answer, category, type, and its corresponding verbose formulation used during question generation. Alternating row shading improves readability.}
    \label{table:examples}
\end{table*}

%% file: supplementary/Z.examples-claims-facts.tex
\begin{table*}
    \centering
    \small
    \resizebox{1\textwidth}{!}{  
    \begin{tabular}{l p{.2\textwidth} p{.3\textwidth} p{.15\textwidth} p{.15\textwidth}}
        \toprule
         & \bf Model Response & \bf Claims & \bf Factuality score & \bf Relevence Score \\
         \midrule
\textbf{Claude 3.5 Sonnet}  & They skin the Father and make a mask and breastplate out of his treated flesh & \colorbox{BurntOrange!40}{(1)}  The group skins the Father.  \colorbox{blue!40}{(2)}  The group makes a mask out of the treated flesh of the Father.  \colorbox{red!30}{(3)}  The group makes a breastplate out of the treated flesh of the Father. & \colorbox{BurntOrange!40}{(1)} Supported \colorbox{blue!40}{(2)} Supported \colorbox{red!30}{(3)} Supported \textit{  \bf (Overall 5)} & \colorbox{BurntOrange!40}{(1)} Relevant \colorbox{blue!40}{(2)} Relevant \colorbox{red!30}{(3)} Relevant \textit{  \bf (Overall 5)} \\ \\
\textbf{Gemini-2.5-Flash}  & The group member Sweetie calls her father from a phone booth. Based on this call, her father, a retired police officer, contacts the police and provides them with the location of the museum where his daughter called from. & \colorbox{BurntOrange!40}{(1)}  Sweetie calls her father from a phone booth.  \colorbox{blue!40}{(2)}  Sweetie's father is a retired police officer.  \colorbox{red!30}{(3)}  Sweetie's father contacts the police.  \colorbox{yellow!80}{(4)}  Sweetie's father provides the police with the location of the museum.  \colorbox{green!40}{(5)}  The call from Sweetie occurs at the phone booth.  \colorbox{blue!20}{(6)}  The museum is the location from which Sweetie called her father. & \colorbox{BurntOrange!40}{(1)} Unsupported \colorbox{blue!40}{(2)} Unsupported \colorbox{red!30}{(3)} Unsupported \colorbox{yellow!80}{(4)} Unsupported \colorbox{green!40}{(5)} Unsupported \colorbox{blue!20}{(6)} Unsupported \textit{  \bf (Overall 0)} & \colorbox{BurntOrange!40}{(1)} NotRelevant \colorbox{blue!40}{(2)} NotRelevant \colorbox{red!30}{(3)} NotRelevant \colorbox{yellow!80}{(4)} NotRelevant \colorbox{green!40}{(5)} NotRelevant \colorbox{blue!20}{(6)} NotRelevant \textit{  \bf (Overall 2)} \\ \\ 
\textbf{GPT-4o}  & They skin Father and make a mask and breastplate out of his treated flesh. & \colorbox{BurntOrange!40}{(1)}  The group skins Father.  \colorbox{blue!40}{(2)}  The group makes a mask out of Father's treated flesh.  \colorbox{red!30}{(3)}  The group makes a breastplate out of Father's treated flesh. & \colorbox{BurntOrange!40}{(1)} Supported \colorbox{blue!40}{(2)} Supported \colorbox{red!30}{(3)} Supported \textit{  \bf (Overall 5)} & \colorbox{BurntOrange!40}{(1)} Relevant \colorbox{blue!40}{(2)} Relevant \colorbox{red!30}{(3)} Relevant \textit{  \bf (Overall 5)} \\ \\ 
\textbf{Amazon Nova Lite}  & The group burns the mask that the father used. & \colorbox{BurntOrange!40}{(1)}  The group burns the mask.  \colorbox{blue!40}{(2)}  The mask was used by the father. & \colorbox{BurntOrange!40}{(1)} Unsupported \colorbox{blue!40}{(2)} Unsupported \textit{  \bf (Overall 1)} & \colorbox{BurntOrange!40}{(1)} NotRelevant \colorbox{blue!40}{(2)} NotRelevant \textit{  \bf (Overall 2)} \\ \\ 
\textbf{LLaVA-NeXT-Video}  &  The group takes the father's ashes and uses them to make a statue of him. & \colorbox{BurntOrange!40}{(1)}  The group takes the father's ashes.  \colorbox{blue!40}{(2)}  The group uses the father's ashes to make a statue of him. & \colorbox{BurntOrange!40}{(1)} Unsupported \colorbox{blue!40}{(2)} Unsupported \textit{  \bf (Overall 1)} & \colorbox{BurntOrange!40}{(1)} Partially relevant \colorbox{blue!40}{(2)} NotRelevant \textit{  \bf (Overall 2)} \\ \\ 
\textbf{Qwen2.5VL}  &  The group makes a mask and breastplate out of the father's treated flesh. & \colorbox{BurntOrange!40}{(1)}  The group makes a mask out of the father's treated flesh.  \colorbox{blue!40}{(2)}  The group makes a breastplate out of the father's treated flesh. & \colorbox{BurntOrange!40}{(1)} Supported \colorbox{blue!40}{(2)} Supported \textit{  \bf (Overall 5)} & \colorbox{BurntOrange!40}{(1)} Relevant \colorbox{blue!40}{(2)} Relevant \textit{  \bf (Overall 5)} \\ \\ 
\textbf{Mini-CPM-o}  &  The group skins the father and makes a mask and breastplate out of his treated flesh.
 & \colorbox{BurntOrange!40}{(1)}  The group skins the father.  \colorbox{blue!40}{(2)}  The group makes a mask out of the father's treated flesh.  \colorbox{red!30}{(3)}  The group makes a breastplate out of the father's treated flesh. & \colorbox{BurntOrange!40}{(1)} Supported \colorbox{blue!40}{(2)} Supported \colorbox{red!30}{(3)} Supported \textit{  \bf (Overall 5)} & \colorbox{BurntOrange!40}{(1)} Relevant \colorbox{blue!40}{(2)} Relevant \colorbox{red!30}{(3)} Relevant \textit{  \bf (Overall 5)} \\ \\ 
         \bottomrule
    \end{tabular}
    }
    \caption{Representative examples for models responses and their evaluation on the question \textbf{Q.}\textbf{ What does the group do with something they have from the father?} \emph{The group skins the father and makes a mask and breastplate out of his treated flesh.}}
    \label{table:examples-claims}
\end{table*}


%% file: supplementary/Z.all_prompts.tex
\begin{table*}
    \centering
    \small
    \resizebox{.93\textwidth}{!}{  
    \begin{tabular}{p{.9\textwidth}}
        \toprule
        \textbf{Instructions}: You are an expert factuality evaluator for video question answering systems. Your task is to evaluate the factual accuracy of claims made in a model’s response by comparing them against ground truth atomic facts and dialogue from the same video segment. \\
        \midrule

        \textbf{Input Information:} \\
        \textit{Question}: \{question\} \\
        \textit{Claims Extracted from Model Response}: \\
        \{claims\} \\
        \textit{Ground Truth Atomic Facts from Video Segment}: \\
        \{facts\} \\
        \textit{SRT Dialogue Context}: \\
        \{context\} \\ \\

        \textbf{Evaluation Task:} \\
        For each claim in the model response, evaluate its factual accuracy by checking against: \\
        1. The ground truth atomic facts \\
        2. The SRT dialogue context \\ \\

        \textbf{Scoring Rubric:} For each claim, assign one label: \\
        \quad -- \textit{SUPPORTED (S)}: Directly supported by the facts or dialogue. \\
        \quad -- \textit{PARTIALLY\_SUPPORTED (PS)}: Some accurate information, but imprecise or partially incorrect. \\
        \quad -- \textit{UNSUPPORTED (U)}: Not supported by any fact or dialogue. \\
        \quad -- \textit{CONTRADICTORY (C)}: Directly contradicts facts or dialogue. \\
        \quad -- \textit{NOT\_CHECKABLE (NC)}: Cannot be verified from the provided sources. \\ \\

        \textbf{Important Guidelines:} \\
        1. Penalize hallucinations strictly. Unsupported additions should be marked U or C. \\
        2. Consider the dialogue when verifying the claims. It may contain information not in the atomic facts. \\ \\

        \textbf{Factuality Score (0--5):} \\
        \textit{5}: All claims supported; no hallucinations. \\
        \textit{4}: Mostly supported; minor partial issues. \\
        \textit{3}: Mix of supported and partial; some unsupported but no contradictions. \\
        \textit{2}: Multiple unsupported claims or one contradiction. \\
        \textit{1}: Mostly unsupported or contradictory. \\
        \textit{0}: Entirely incorrect, unsupported, or contradictory. \\ \\

        \textbf{Examples:} \\
        \textit{Supported}: Claim matches facts and dialogue. \\
        \textit{Partially Supported}: Claim has some accuracy but adds imprecise details. \\
        \textit{Unsupported}: Claim adds unverifiable information. \\
        \textit{Contradictory}: Claim conflicts with facts/dialogue. \\ \\

        \textbf{Overall Score Examples:} \\
        Score 5: All claims supported. \\
        Score 4: Mostly supported with one minor issue. \\
        Score 3: Mix of supported/partial. \\
        Score 2: Includes unsupported or contradictory. \\
        Score 1: Mostly hallucinated. \\
        Score 0: Fully wrong or contradictory. \\ \\

        At the end, evaluate all claims according to these guidelines. \\
        
        \bottomrule
    \end{tabular}
    }
    \caption{\textbf{Prompt for Reference-Free Factuality Evaluation.}}
    \label{table:prompt_factuality_evaluation}
\end{table*}

\begin{table*}
    \centering
    \small
    \resizebox{\textwidth}{!}{
    \begin{tabular}{p{\textwidth}}
        \toprule
        \textbf{Instructions}: You are an expert relevance evaluator for video question answering systems. Your task is to evaluate the relevance of claims made in a model’s response by comparing them against the user’s question. The ground truth facts and dialogue should be used only as context to understand the claims, not to judge correctness. \\
        \midrule

        \textbf{Input Information:} \\
        \textit{Question}: \{question\} \\
        \textit{Claims Extracted from Model Response}: \\
        \{claims\} \\
        \textit{Ground Truth Atomic Facts Used to Answer the Question}: \\
        \{facts\} \\
        \textit{SRT Dialogue Context}: \\
        \{context\} \\ \\

        \textbf{Evaluation Task:} \\
        For each claim, evaluate its relevance to the question. \\
        1. Compare the claim’s topic to the question’s topic. \\
        2. Use the facts and dialogue to understand the meaning of the claim. \\
        3. Do \textbf{not} evaluate factual correctness. A claim may be wrong but still relevant. \\ \\

        \textbf{Scoring Rubric:} \\
        \textit{RELEVANT (R)}: Directly answers or is clearly pertinent to the question. \\
        \textit{PARTIALLY\_RELEVANT (PR)}: Related to the general topic but tangential or not directly asked for. \\
        \textit{NOT\_RELEVANT (NR)}: Off-topic; does not help answer the question. \\ \\

        \textbf{Important Guidelines:} \\
        Relevance is independent of correctness. \\
        Claims answering only part of a multi-part question are still relevant. \\
        Tangential information should be marked PR or NR. \\ \\

        \textbf{Relevance Score (0--5):} \\
        5: All claims are RELEVANT. Perfect focus. \\
        4: Mostly RELEVANT; at most minor PR tangents. \\
        3: Mix of R and PR; no NR claims. \\
        2: At least one NR claim; answer loses focus. \\
        1: Multiple NR claims; mostly off-topic. \\
        0: All claims NR; completely irrelevant. \\ \\

        \textbf{Examples:} \\
        \textit{Relevant}: Answers “who” and “what he does.” \\
        \textit{Partially Relevant}: Describes the correct entity but not the attribute asked for. \\
        \textit{Not Relevant}: About a different person, not answering the question. \\
        \textit{Relevant but Incorrect}: Still relevant if it tries to answer the question. \\ \\

        After evaluating all claims, assign a final relevance score following these rules. \\

        \bottomrule
    \end{tabular}
    }
    \caption{\textbf{Prompt for Reference-Free Relevance Evaluation.}}
    \label{table:prompt_relevance_evaluation}
\end{table*}

\begin{table*}
    \centering
    \small
    \resizebox{\textwidth}{!}{
    \begin{tabular}{p{\textwidth}}
        \toprule
        \textbf{Instructions}: You are an expert coherence evaluator for video question answering systems. Your task is to evaluate the internal logical coherence of a model’s response by comparing the claims it makes against one another. A coherent response should not contain contradictions or excessive redundancy. \\
        \midrule

        \textbf{Input Information:} \\
        \textit{Question}: \{question\} \\
        \textit{Claims Extracted from Model Response}: \\
        \{claims\} \\ \\

        \textbf{Evaluation Task:} \\
        For each claim, evaluate its logical consistency with all other claims in the same response. The goal is to detect internal contradictions or redundancies; external factual correctness should \textbf{not} be considered. \\ \\

        \textbf{Scoring Rubric:} \\
        \textit{CONSISTENT (CO)}: Logically consistent with all other claims. Introduces new information without conflict. \\
        \textit{REDUNDANT (R)}: Repeats information expressed in another claim using different phrasing. A minor coherence flaw. \\
        \textit{CONTRADICTORY (C)}: Directly contradicts one or more other claims. \\ \\

        \textbf{Important Guidelines:} \\
        -- Look for direct logical opposites (e.g., “Tony is happy” vs. “Tony is sad”). \\
        -- Increased specificity is \textbf{not} contradiction (e.g., “doctor” vs. “brain surgeon”). \\
        -- Redundancy occurs when a claim adds no new information. \\
        -- For REDUNDANT or CONTRADICTORY labels, reference the related claim number in the justification. \\ \\

        \textbf{Coherence Score (0--5):} \\
        5: All claims CONSISTENT; no redundancy or contradictions. \\
        4: Mostly CONSISTENT; at most one or two REDUNDANT claims; no contradictions. \\
        3: Several REDUNDANT claims but no CONTRADICTORY claims. \\
        2: At least one CONTRADICTORY pair. \\
        1: Multiple CONTRADICTORY claims; logically confusing. \\
        0: Most claims CONTRADICTORY; response is incoherent. \\ \\

        \textbf{Examples:} \\
        CONSISTENT: Introduces new, non-conflicting information. \\
        REDUNDANT: Restates ideas already given (e.g., “in charge of the hospital” vs. “head of the hospital”). \\
        CONTRADICTORY: Directly conflicts with previous claims (e.g., “head of the hospital” vs. “junior intern”). \\ \\

        After evaluating all claims, assign a final coherence score following these rules. \\

        \bottomrule
    \end{tabular}
    }
    \caption{\textbf{Prompt for Reference-Free Coherence Evaluation.}}
    \label{table:prompt_coherence_evaluation}
\end{table*}